%% file: main.tex
\documentclass{article} %
\usepackage{iclr2024_conference,times}
\iclrfinalcopy

\input{packages}
\input{math_commands.tex}
\renewcommand{\eqref}[1]{Eq.~(\ref{#1})}

\title{Prompt-tuning latent diffusion models\\ for inverse problems}

\author{Hyungjin Chung\thanks{This work was done during an internship at Google}$^{\,\,\,\,1,2}$, Jong Chul Ye$^{2}$, Peyman Milanfar$^{1}$ \& Mauricio Delbracio$^{1}$ \\
$^{1}$ Google Research, \quad $^{2}$ KAIST\\
\texttt{\{hj.chung, jong.ye\}@kaist.ac.kr}, \quad \texttt{\{milanfar, mdelbra\}@google.com}
}

\begin{document}

\maketitle

\begin{abstract}
We propose a new method for solving imaging inverse problems using text-to-image latent diffusion models as general priors. Existing methods using latent diffusion models for inverse problems typically rely on simple null text prompts, which can lead to suboptimal performance. To address this limitation, we introduce a method for prompt tuning, which jointly optimizes the text embedding on-the-fly while running the reverse diffusion process. This allows us to generate images that are more faithful to the diffusion prior. In addition, we propose a method to keep the evolution of latent variables within the range space of the encoder, by projection. This helps to reduce image artifacts, a major problem when using latent diffusion models instead of pixel-based diffusion models. Our combined method, called P2L, outperforms both image- and latent-diffusion model-based inverse problem solvers on a variety of tasks, such as super-resolution, deblurring, and inpainting.
\end{abstract}

\section{Introduction}
\label{sec:intro}
Imaging inverse problems are often solved by optimizing or sampling  a functional that combines a data-fidelity/likelihood term with a regularization term or signal prior~\citep{romano2017little,venkatakrishnan2013plug,ongie2020deep,kamilov2023plug, kawar2022denoising,kadkhodaie2020solving,chung2023diffusion}. A common regularization strategy is to use pre-trained image priors from generative models, such as GANs
~\citep{bora2017compressed}, VAEs~\citep{bora2017compressed,gonzalez2022solving}, Normalizing flows~\citep{whang2021solving} or Diffusion models~\citep{song2022solving,chung2022score}.

In particular, diffusion models have gained significant attention as implicit generative priors for solving inverse problems in imaging~\citep{kadkhodaie2020solving,whang2022deblurring,daras_dagan_2022score,kawar2022denoising,feng2023score,laroche2023fast,chung2023diffusion}. Leaving the pre-trained diffusion prior intact, one can guide the inference process to perform posterior sampling conditioned on the measurement at inference time by resorting to Bayesian inference. In the end, the ultimate goal of Diffusion model-based Inverse problem Solvers (DIS) would be to act as a fully general inverse problem solver, which can be used not only regardless of the imaging model, but also regardless of the data distribution. 

Solving inverse problems in a fully general domain is hard. This directly stems from the difficulty of generative modeling a wide distribution, where it is known that one has to trade-off diversity with fidelity by some means of sharpening the distribution~\citep{brock2018large,dhariwal2021diffusion}. The standard approach in modern diffusion models is to condition on text prompts~\citep{rombach2022high,saharia2022photorealistic}, among them the most popular being Stable Diffusion (SD), a latent diffusion model (LDM), which is itself an under-explored topic in the context of inverse problem solving. While text conditioning is now considered standard practice in content creation including images~\citep{ramesh2022hierarchical,saharia2022photorealistic}, 3D~\citep{poole2023dreamfusion,wang2023prolificdreamer}, video~\citep{ho2022imagen}, personalization~\citep{gal2022image}, and editing~\citep{hertz2022prompt}, it has been completely disregarded in the inverse problem solving context. This is natural, as it is highly ambiguous which text would be beneficial to use when all we have is a degraded measurement. The wrong prompt could easily lead to degraded performance.

In this work, we aim to bridge this gap by proposing a way to {\em automatically} find the right prompt to condition diffusion models when solving inverse problems. This can be achieved through optimizing the continuous text embedding {\em on-the-fly} while running DIS. We formulate this into a Bayesian framework of updating the text embedding and the latent in an alternating fashion, such that they become gradually aligned during the sampling process. 
Orthogonal and complementary to embedding optimization, we devise a simple LDM-based DIS (LDIS) that controls the evolution of the latents to stay on the natural data manifold by explicit projection.
We name the algorithm that combines these components P2L, short for \textbf{P}rompt-tuning \textbf{P}rojected \textbf{L}atent diffusion model-based inverse problem solver.
In reaching for the ultimate goal of DIS, we focus on 1) \textbf{LDM-based DIS} (LDIS) for solving inverse problems in the 2) \textbf{fully general domain} (using a single pre-trained checkpoint) that targets 3) \textbf{512$\times$512 resolution}\footnote{All prior works on DIS/LDIS focused on 256$\times$256 resolution. Most LDIS focused their evaluation on a constrained dataset such as FFHQ, and did not scale their method to more general domains such as ImageNet.}. All the aforementioned components are highly challenging, and to the best of our knowledge, have not been studied in conjunction before.

\begin{table}[t]
\setlength{\tabcolsep}{2pt}
\centering
\resizebox{1.0\textwidth}{!}{%
\begin{tabular}{lcccccccccccc}
\toprule
{} & \multicolumn{6}{c}{\textbf{FFHQ}} & \multicolumn{6}{c}{\textbf{ImageNet}} \\
\cmidrule(lr){2-7}
\cmidrule(lr){8-13}
{} & \multicolumn{3}{c}{SR$\times 8$} & \multicolumn{3}{c}{Inpaint ($p=0.8$)} & \multicolumn{3}{c}{SR$\times 8$} & \multicolumn{3}{c}{Inpaint ($p=0.8$)} \\
\cmidrule(lr){2-4}
\cmidrule(lr){5-7}
\cmidrule(lr){8-10}
\cmidrule(lr){11-13}
{\textbf{Prompt}} & {FID$\downarrow$} & {LPIPS$\downarrow$} & {PSNR$\uparrow$} & {FID$\downarrow$} & {LPIPS$\downarrow$} & {PSNR$\uparrow$} & {FID$\downarrow$} & {LPIPS$\downarrow$} & {PSNR$\uparrow$} & {FID$\downarrow$} & {LPIPS$\downarrow$} & {PSNR$\uparrow$}\\
\midrule
\texttt{""} & 61.16 & 0.327 & 26.49 & 52.34 & 0.241 & \textbf{29.78} & 78.68 & 0.397 & 23.49 & 70.87 & 0.350 & \underline{26.20}\\
\cmidrule{1-13}
\texttt{"A high quality photo"} &  61.17 & 0.327 & 26.57 & 52.82 & \underline{0.237} & 29.70 & 77.00 & 0.396 & 23.51 & 69.10 & 0.350 & \underline{26.26} \\
\texttt{"A high quality photo of a cat"} &  69.03 & 0.377 & 26.39 & 55.15 & 0.248 & 29.63 & 76.69 & 0.402 & \textbf{23.63} & 68.48 & 0.355 & 26.13 \\
\texttt{"A high quality photo of a dog"} &  66.55 & 0.371 & 26.48 & 55.91 & 0.249 & 29.65 & 76.45 & 0.394 & 23.58 & 67.75 & 0.354 & 26.10 \\
\texttt{"A high quality photo of a face"} &  \underline{60.41} & 0.325 & 26.74 & 52.33 & 0.239 & 29.69 & 77.32 & 0.403 & \underline{23.60} & 68.83 & 0.352 & \underline{26.20} \\
\cmidrule{1-13}
\rowcolor{BrickRed!10}
Proposed &  \textbf{58.73} & \textbf{0.317} & 26.68 & \textbf{51.40} & \textbf{0.233} & 29.69 & \underline{66.96} & \textbf{0.386} & 23.57 & \underline{66.82} & \textbf{0.314} & \textbf{26.29} \\
\cmidrule{1-13}
\rowcolor{mylightblue}
PALI prompts from $\y$ &  61.33 & 0.329 & \textbf{26.81} & 54.34 & 0.249 & \underline{29.76} & 68.28 & 0.388 & 23.57 & 69.55 & 0.355 & \underline{26.26} \\
\rowcolor{mylightblue}
PALI prompts from $\x$ &  60.73 & \underline{0.322} & \underline{26.76} & \underline{52.06} & 0.238 & 29.75 & \textbf{66.55} & \underline{0.387} & 23.57 & \textbf{64.00} & \underline{0.348} & 26.17 \\
\bottomrule
\end{tabular}
}
\caption{
Difference in restoration performance using LDPS on SR$\times 8$ task with varying text prompts. Proposed: text embedding optimized without access to ground truth. PALI prompts from $\x$/$\y$: captions are generated with PALI~\citep{chen2022pali} from $\x$: ground truth clean images / $\y$: degraded images. The former can be considered an empirical upper bound.
}
\label{tab:proof_of_concept_ptip}
\end{table}

\section{Background}
\label{sec:background}

\subsection{Latent diffusion models}
\label{sec:latent_diffusion_models}

Diffusion models are generative models that learn to reverse the forward noising process~\citep{sohl2015deep,ho2020denoising,song2020score}, starting from the initial distribution $p_0(\x),\, \x \in \Rd^n$ and approaching the standard Gaussian $p_T(\x) = \Nc(\bm{0}, \Ib)$ as $T \rightarrow \infty$. Considering the variance-preserving (VP) formulation~\citep{ho2020denoising}, the forward/reverse processes can be characterized with Ito stochastic differential equations (SDE)~\citep{song2020score}
\begin{align}
\label{eq:forward_sde}
    d\x_t &= -\frac{\beta_t}{2}\x_t dt + \sqrt{\beta_t}d\w && \text{(Forward)} \\
    d\x_t &= \left[-\frac{\beta_t}{2}\x - \beta_t\nabla_{\x_t} \log p_t({\x_t})\right]dt + \sqrt{\beta_t}d\bar\w && \text{(Reverse)},
\label{eq:reverse_sde}
\end{align}
where $\beta_t$ is the noise schedule\footnote{We adopt standard notations for the noise schedule $\beta_t, \alpha_t, \bar\alpha_t$ from~\cite{ho2020denoising}.} and $\w, \bar\w$  are the standard forward/reverse Wiener processes.
Here $\nabla_{\x_t} \log p_t(\x_t)$ is typically approximated with a score network $\s_\thetab(\cdot)$ or a noise estimation network $\epsilonb_\thetab(\cdot)$, and learned through denoising score matching (DSM)~\citep{vincent2011connection} or epsilon-matching loss~\citep{ho2020denoising}. 

Image diffusion models that operate on the pixel space $\x$ are compute-heavy. One workaround for compute-efficient generative modeling is to leverage an autoencoder~\citep{rombach2022high,kingma2013auto}
\begin{align}
\label{eq:ldm_vae}
    \Ec: \Rd^n \mapsto \Rd^k,\, \Dc: \Rd^k \mapsto \Rd^n,\, \x \simeq \Dc(\Ec(\x)) \quad \forall \x \sim p_{\rm data}(\x),
\end{align}
where $\Ec$ is the encoder, $\Dc$ is the decoder, and $k < n$. After encoding the images into the {\em latent} space $\z = \Ec(\x)$~\citep{rombach2022high}, one can train a diffusion model in the low-dimensional latent space. Latent diffusion models (LDM) are beneficial in that the computation is cheaper as it operates in a lower-dimensional space, consequently being more suitable for modeling higher dimensional data (e.g. large images of size $\geq 512^2$). The effectiveness of LDMs have democratized the use of diffusion models as the de facto standard of generative models especially for images under the name of Stable Diffusion (SD), which we focus on extensively in this work.

One notable difference of SD from standard image diffusion models~\citep{dhariwal2021diffusion} is the use of text conditioning $\epsilonb_\thetab(\cdot, \Cc)$, where $\Cc$ is the continuous embedding vector usually obtained through the CLIP text embedder~\citep{radford2021learning}. As the model is trained with LAION-5B~\citep{schuhmann2022laion}, a large-scale dataset containing image-text pairs, SD can be conditioned during the inference time to generate images that are aligned with the given text prompt by directly using $\epsilonb_\thetab(\cdot, \Cc)$, or by means of classifier-free guidance (CFG)~\citep{ho2021classifierfree}.

\subsection{Solving inverse problem with (latent) diffusion models}
\label{sec:ldis}

Given access to some measurement
\begin{align}
    \y = \Ab\x + \n,\quad \x \in \Rd^n,\, \y \in \Rd^m,\, \Ab \in \Rd^{m \times n},\, \n \sim \Nc(\bm{0}, \sigma_y^2\Ib_m)
\end{align}
where $\Ab$ is the forward operator and $\n$ is additive white Gaussian noise, the task is retrieving $\x$ from $\y$. As the problem is ill-posed, a natural way to solve it is to perform posterior sampling $\x \sim p(\x|\y)$ by defining a suitable prior $p(\x)$. In DIS, diffusion models (i.e. denoisers) act as the implicit prior with the use of the score function.

Earlier methods utilized an alternating projection approach, where hard measurement constraints are applied in-between the denoising steps whether in pixel space~\citep{kadkhodaie2020solving,song2020score} or measurement space~\citep{song2022solving,chung2022score}. Distinctively, projection in the spectral space via singular value decomposition (SVD) to incorporate measurement noise has been developed~\citep{kawar2021snips,kawar2022denoising}. Subsequently, methods that aim to approximate the gradient of the log posterior in the diffusion model context have been proposed~\citep{chung2023diffusion,song2023pseudoinverseguided}, expanding the applicability to nonlinear problems. Broadening the range even further, methods that aim to solve blind~\citep{chung2023parallel,murata2023gibbsddrm}, 3D~\citep{chung2023solving,lee2023improving}, and unlimited resolution problems~\citep{wang2023unlimited} were introduced. More recently, methods leveraging diffusion score functions within variational inference to solve inverse imaging has been proposed~\citep{mardani2023variational,feng2023score}. Notably, all the aforementioned methods utilize {\em image-domain} diffusion models. Orthogonal to this direction, some of the recent works have shifted their attention to using {\em latent} diffusion models~\citep{rout2023solving,song2023solving,he2023iterative}, a direction that we follow in this work.

One canonical DIS that covers the widest range of non-blind problems is diffusion posterior sampling (DPS)~\citep{chung2023diffusion}, which proposes to approximate
\begin{align}
    \nabla_{\x_t} \log p(\y|\x_t) \simeq \nabla_{\x_t} \log p(\y|\Ed[\x_0|\x_t]),\quad
    \Ed[\x_0|\x_t] = \frac{1}{\sqrt{\bar\alpha_t}}\left(\x_t - \sqrt{1 - \bar\alpha_t}\epsilonb_{\thetab^*}(\x_t)\right),
\end{align}
where the posterior mean is the result of Tweedie's formula~\citep{robbins1956empirical,efron2011tweedie,chung2023diffusion}.
This idea was recently extended to LDMs in a few recent works~\citep{rout2023solving,he2023iterative}
\begin{align}
\label{eq:ldps}
    \nabla_{\z_t} \log p(\y|\z_t) \simeq \nabla_{\z_t} \log p(\y|\Dc(\Ed[\z_0|\z_t])) = \nabla_{\z_t} \|\y - \Dc(\hat\z_0)\|_2^2,
\end{align}
with $\hat\z_0 := \Ed[\z_0|\z_t]$. We refer to the sampler that uses the approximation in \eqref{eq:ldps} as Latent DPS (LDPS) henceforth. \cite{rout2023solving} extends LDPS with an additional regularization term to guide the latent to be the fixed point of the autoencoding process, and \cite{he2023iterative} extends LDPS by using history updates as in Adam~\citep{kingma2015adam}. However, {\em all} of the existing works in the literature that aims for LDIS, to the best of our knowledge, neglects the use of text embedding by resorting to the use of null text embedding $\Cc_\varnothing$.

\subsection{Prompt tuning}
\label{sec:pt}

In modern language models and vision-langauge models, {\em prompting} is a standard technique~\citep{radford2021learning,brown2020language} to guide the large pre-trained models to solve downstream tasks. As it has been found that even slight variations in the prompting technique can lead to vastly different outcomes~\citep{kojima2022large}, prompt tuning (learning) has been introduced~\citep{shin2020autoprompt,zhou2022learning}, which defines a {\em learnable} context vector to optimize over. It was shown that by only optimizing over the continuous embedding vector while maintaining the model parameters fixed, one can achieve a significant performance gain.

In the context of diffusion models, prompt tuning has been adopted for personalization~\citep{gal2022image}, where one defines a special token to embed a specific concept with only a few images. Moreover, it has also been demonstrated that one can achieve superior editing performance by optimizing for the null text prompt $\Cc_{\varnothing}$~\citep{mokady2023null} before the reverse diffusion sampling process.

\section{Main Contribution: the P2L algorithm}
\label{sec:main}

\begin{figure}[t]
\begin{minipage}{.49\textwidth}
    \begin{algorithm}[H]
        \small
        \caption{Update $\Cc_t$}
        \label{alg:fn:optimize_emb}
        \begin{algorithmic}[1]
        \Function{$\color{xkcdWine}{\text{OptimizeEmb}}$}{$\z_t, \y, \Cc_t^{(0)}$}
            \For{$k=1$ {\bfseries to} $K$}
                \State $\hat\epsilonb_{t} \gets \epsilonb_{\thetab^{*}}(\z_{t}, \Cc_t^{(k-1)})$
                \State $\hat\z_{0|t} \gets (\z_t - \sqrt{1 - \bar\alpha_t}
                \hat{\epsilonb}_{t}){/\sqrt{\bar\alpha_t}}$
                \State $\hat\x_{0|t} \gets \Dc(\hat\z_{0|t})$
                \State $\Lc \gets \|\Ab\hat\x_{0|t}(\Cc_t^{(k-1)}) - \y\|_2^2$
                \State $\Cc_t^{(k)} \gets \Cc_t^{(k-1)} - \texttt{AdamGrad}(\Lc_t)$
            \EndFor
            \State \Return $\Cc_t^* \gets \Cc_t^{(K)}$
        \EndFunction
        \end{algorithmic}
    \end{algorithm}
\end{minipage}
\begin{minipage}{.49\textwidth}
    \begin{algorithm}[H]
        \small
        \caption{Update $\z_t$}
        \label{alg:p2l}
        \begin{algorithmic}[1]
        \Require $\epsilonb_{\thetab^{*}}, \z_T, \y, \Cc, T, K$
        \For{$t=T$ {\bfseries to} $1$}
            \State $\Cc_t^{*} \gets \color{xkcdWine}{\textsc{OptimizeEmb}}(\z_t, \y, \Cc_t^{0})$
            \State $\hat\epsilonb_{t} \gets \epsilonb_{\thetab^{*}}(\z_{t}, \Cc_t^{*})$
            \State $\hat\z_{0|t} \gets (\z_t - \sqrt{1 - \bar\alpha_t} \hat{\epsilonb}_{t}){/\sqrt{\bar\alpha_t}}$
            \State $\color{xkcdOrange}{\hat\z'_{0|t} \gets \Ec\left({\bm\Gamma}\left(\Dc(\hat\z_{0|t})\right)\right)}$
            \State $\z'_{t-1} \gets \sqrt{\bar\alpha_{t-1}}\hat\z_{0|t} + \sqrt{1 - \bar\alpha_{t-1}}\hat\epsilonb_t$
            \State $\z_{t-1} \gets \z'_{t-1} - \rho_{t}\nabla_{\z_t}\|\Ac\Dc(\hat\z_{0|t}) - \y\|$
            \State $\Cc_{t-1}^{(0)} \gets \Cc_t^*$
            \EndFor
        \State {\bfseries return} $\x_0 \gets \Dc(\z_0)$
        \end{algorithmic}
    \end{algorithm}
\end{minipage}
\end{figure}

\begin{figure}[t]
\begin{minipage}{.49\textwidth}
    \begin{algorithm}[H]
        \small
        \caption{Prompt tuning}
        \label{alg:fn:optimize_emb}
        \begin{algorithmic}[1]
        \Function{$\color{xkcdWine}{\text{OptimizeEmb}}$}{$\z_t, \y, \Cc_t^{(0)}$}
            \For{$k=1$ {\bfseries to} $K$}
                \State $\hat\epsilonb_{t} \gets \epsilonb_{\thetab^{*}}(\z_{t}, \Cc_t^{(k-1)})$
                \State $\hat\z_{0|t} \gets (\z_t - \sqrt{1 - \bar\alpha_t}
                \hat{\epsilonb}_{t}){/\sqrt{\bar\alpha_t}}$
                \State $\hat\z'_{0|t} \gets \hat\z_{0|t} - \rho \nabla_{\hat\z_{0|t}}\|\y - \Dc(\hat\z_{0|t})\|$
                \State $\hat\x_{0|t} \gets \Dc(\hat\z'_{0|t})$
                \State $\Lc \gets \|\Ab\hat\x_{0|t}(\Cc_t^{(k-1)}) - \y\|_2^2$
                \State $\Cc_t^{(k)} \gets \Cc_t^{(k-1)} - \texttt{AdamGrad}(\Lc_t)$
            \EndFor
            \State \Return $\Cc_t^* \gets \Cc_t^{(K)}$
        \EndFunction
        \end{algorithmic}
    \end{algorithm}
\end{minipage}
\begin{minipage}{.49\textwidth}
    \begin{algorithm}[H]
        \small
        \caption{P2L}
        \label{alg:p2l}
        \begin{algorithmic}[1]
        \Require $\epsilonb_{\thetab^{*}}, \z_T, \y, \Cc, T, K, \gamma, \bm{\Gamma}$
        \For{$t=T$ {\bfseries to} $1$}
            \State $\Cc_t^{*} \gets \color{xkcdWine}{\textsc{OptimizeEmb}}(\z_t, \y, \Cc_t^{0})$
            \State $\hat\epsilonb_{t} \gets \epsilonb_{\thetab^{*}}(\z_{t}, \Cc_t^{*})$
            \State $\hat\z_{0|t} \gets (\z_t - \sqrt{1 - \bar\alpha_t} \hat{\epsilonb}_{t}){/\sqrt{\bar\alpha_t}}$
            \If {$(t \mod \gamma) = 0$}
                \State $\color{xkcdOrange}{\hat\z'_{0|t} \gets \Ec\left({\bm\Gamma}\left(\Dc(\hat\z_{0|t})\right)\right)}$
            \EndIf
            \State $\z'_{t-1} \gets \sqrt{\bar\alpha_{t-1}}\hat\z'_{0|t} + \sqrt{1 - \bar\alpha_{t-1}}\hat\epsilonb_t$
            \State $\z_{t-1} \gets \z'_{t-1} - \rho_{t}\nabla_{\z_t}\|\Ac\Dc(\hat\z_{0|t}) - \y\|$
            \State $\Cc_{t-1}^{(0)} \gets \Cc_t^*$
            \EndFor
        \State {\bfseries return} $\x_0 \gets \Dc(\z_0)$
        \end{algorithmic}
    \end{algorithm}
\end{minipage}
\end{figure}

\subsection{Prompt tuning inverse problem solver}
\label{sec:ptip}

The objective of solving inverse problems is to provide a restoration that is as close as possible to the ground truth given the measurement, whether we are targeting to minimize the distortion or to maximize the perceptual quality~\citep{blau2018perception, delbracio2023inversion}. Formally, let us denote a loss function $\Lc(\x, \cb)$ that measures the discrepancy from the ground-truth given the estimate of the truth $\x$, and some additional condition $\cb$. In the context of LDIS, we consider
\begin{align}
    \argmin_{\x, \cb} \Lc(\x, \cb) \equiv \argmin_{\z, \cb} \Lc(\Dc(\z), \cb), \quad \x = \Dc(\z),
\end{align}
where $\cb$ is the text embedding, $\Dc$ is the decoder of the VAE, and the loss $\Lc$ can be considered as the negative log posterior in the Bayesian framework. It is easy to see that
\begin{align}
\label{eq:opt_fix}
    \argmin_{\z, \cb} \Lc(\Dc(\z), \cb) \leq \argmin_{\z} \Lc(\Dc(\z), \cb = \Cc_\varnothing),
\end{align}
where $\Cc_\varnothing$ is the text embedding from the null text prompt.
Notably, by keeping one of the variables fixed, we are optimizing for the {\em upper bound} of the objective that we truly wish to optimize over. It would be naturally beneficial to optimize the LHS of~\eqref{eq:opt_fix}, rather than the RHS used in the previous methods.

\paragraph{A motivating example}

To see \eqref{eq:opt_fix} in effect, we conduct two canonical experiments with 256 test images of FFHQ~\citep{karras2019style} and ImageNet~\citep{deng2009imagenet}: super-resolution (SR) of scale $\times 8$ and inpainting with 80\% of the pixels randomly dropped, using the LDPS algorithm. Keeping all the other hyper-parameters fixed, we only vary the text condition for the diffusion model.
In addition to using a general text prompt, we use PALI~\citep{chen2022pali} to provide captions from the ground truth images ($\x$) and from the measurements ($\y$) and use them when running LDPS. Further details on the experiment can be found in Appendix~\ref{sec:proof_of_concept_exp}.
In Table~\ref{tab:proof_of_concept_ptip}, we first see that simply varying the text prompts can lead to dramatic difference the performance. For instance, we see an increase of over 10 FID when we use the text prompts from PALI for the task of $\times 8$ SR on ImageNet. In contrast, using the prompts generated from $\y$ often degrades the performance (e.g. inpainting) as the correct captions cannot be generated.
From this motivating example, it is evident that additionally optimizing for $\cb$ would bring us gains that are orthogonal to the development of the solvers~\citep{rout2023solving,he2023iterative,song2023solving}, a direction that has not been explored in the literature. Indeed, from the table, we see that by applying our prompt tuning approach, we achieve a large performance gain, sometimes even outperforming the PALI captions which has full access to the ground truth when attaining the text embeddings.

\paragraph{Prompt tuning algorithm}
Existing LDIS approaches attempt to sample from $p(\x|\y, \Cc_\varnothing)$, as it is hard to specify a generally good condition $\Cc$ when all we have access to is the corrupted $\y$. Hence, our goal is to find a good $\Cc$ {\em on-the-fly} while solving for the inverse problem. Before diving into the design of the algorithm, let us first revisit \eqref{eq:ldps} for the case where we consider $\Cc$ as a conditioning signal
\begin{align}
    p(\y|\z_t,\Cc) &= \int p(\y|\x_0)p(\x_0|\z_0)p(\z_0|\z_t,\Cc)\,d\x_0\,d\z_0 \\
    &= \Ed_{p(\z_0|\z_t,\Cc)}[p(\y|\Dc(\z_0))] \stackrel{{\rm (DPS)}}{\simeq} p(\y|\Dc(\hat\z_0^{(\Cc)})),
\label{eq:ldps_C}
\end{align}
where the second equality is achieved by setting $p(\x_0|\z_0) = \delta(\x_0 - \Dc(\z_0))$ and the approximation is achieved by pushing the expectation inside similar to DPS~\citep{chung2023diffusion}\footnote{We introduce additional approximation error for LDMs as we additionally have a nonlinear $\Dc$, which is one of the main reasons why scaling DPS naively does not work too well.}, and we define $\hat\z_0^{(\Cc)}:= \Ed[\z_0|\z_t,\Cc] = \frac{1}{\sqrt{\bar\alpha_t}}\left(\z_t + (1 - \bar\alpha_t)\s_{\thetab^*}(\z_t,\Cc)\right)$. Equipped with the approximation in \eqref{eq:ldps_C}, we propose a sampler reminiscent of Gibbs sampling~\citep{geman1984stochastic} to sample from $p(\x_0,\Cc|\y)$, or equivalently $p(\z_0,\Cc|\y)$. Specifically, we alternate between keeping $\z_t$ fixed and sampling from $p(\Cc|\z_t,\y)$, and keeping $\Cc$ fixed and sampling from $p(\z_t|\Cc,\y)$.

\paragraph{Step 1: $\Cc$ update}
This step ensures that $\Cc$ is aligned with the measurement $\y$ and the current diffusion estimate $\z_t$ with the following update
\begin{align}
    \nabla_\Cc \log p(\Cc|\z_t,\y) &= \nabla_\Cc \log p(\y|\z_t, \Cc) + \nabla_\Cc \log p(\Cc|\z_t) \\
    &\simeq \nabla_\Cc \|\Ab\Dc\left(
    \Ed[\z_0|\z_t,\Cc]\right) - \y
    \|_2^2,
\end{align}
where the second line is achieved by leveraging \eqref{eq:ldps_C}, placing a uniform prior on $p(\Cc)$, and assuming the independence between $\Cc$ and $\z_t$.
In practice, we find that using a few iterative updates with stochastic optimizers such as Adam~\citep{kingma2015adam} is useful. Further, using the conditional posterior mean $\Ed[\z_0|\z_t,\Cc, \y]$ instead of the unconditional posterior mean $\Ed[\z_0|\z_t, \Cc]$, which can be effectively achieved by shifting the posterior mean with a gradient update step~\citep{ravula2023optimizing,barbano2023steerable}, slightly improved performance. 

\paragraph{Step 2: $\z_t$ update}
Let us denote $\Cc_t^*$ the optimized text embedding found through optimization in Step 1 for step $t$. Then, the update step for $\z_t$ reads
\begin{align}
    \nabla_{\z_t} \log p(\z_t|\y, \Cc_t^*) &= \nabla_{\z_t} \log p(\z_t| \Cc_t^*) + \nabla_{\z_t} \log p(\y|\z_t,\Cc_t^*) \\
    &\simeq \s_{\thetab^*}(\z_t, \Cc_t^*) + \rho_t \nabla_{\z_t} \|\Ab\Dc(\Ed[\z_0|\z_t,\Cc_t^*]) - \y\|_2^2,
    \label{eq:ldps_C_opt}
\end{align}
where we used $\nabla_{\z_t} \log p(\z_t|\Cc_t^*) \simeq \s_{\thetab^*}(\z_t, \Cc_t^*)$\footnote{Only using eq.~\ref{eq:ldps_C_opt} with $\Cc_t^* = \Cc_\varnothing$ would result in standard LDPS.}, and we set $\rho_t$ to be the step size that weights the likelihood, similar to~\cite{chung2023diffusion}. We summarize our alternating sampling method in Algorithm~\ref{alg:fn:optimize_emb},\ref{alg:p2l}.
Further details on the implementation and the choice of hyper-parameters can be found in Appendix~\ref{sec:imp_details}.

\subsection{Projection to the range space of $\Ec$}
\label{sec:projection_encoder}

Recall that for both updates proposed in Section.~\ref{sec:ptip}, we introduce an approximation $p(\y|\z_t, \Cc) \simeq p(\y|\Dc(\hat\z_0))$. Here, the decoder introduces a significant amount of error especially when the estimated clean latent $\hat\z_0$ falls of the manifold of the clean latents, which inevitably happens with the LDPS approximation. \cite{rout2023solving} proposed to regularize the update steps on the latent so that the clean latents are led to the fixed point of the successive application of decoding-encoding. Formally, they use the following gradient step
\begin{align}
\label{eq:psld_approx}
    \nabla_{\z_t} \log (\y|\z_t) \simeq \nabla_{\z_t} \left( \|\y - \Dc(\hat\z_0)\|_2^2 + \lambda \|\hat\z_0 - \Ec(\Dc(\hat\z_0))\|_2^2 \right),
\end{align}

\begin{wrapfigure}[]{r}{0.5\textwidth}
\includegraphics[width=0.45\textwidth]{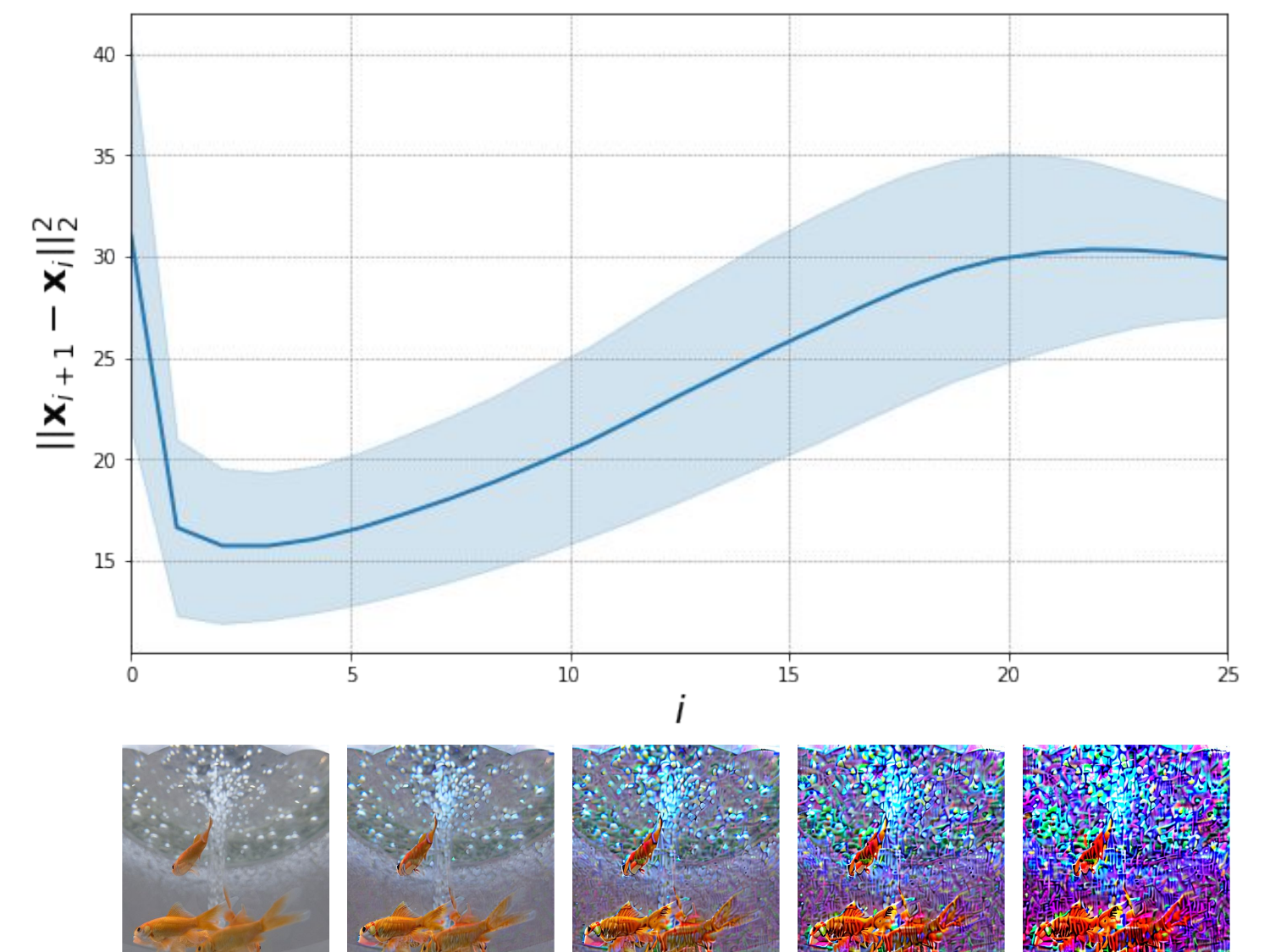}
\caption{Fixed point analysis: $\mu \pm \sigma$}
\label{fig:fixed_point}
\end{wrapfigure}

where the additional regularization term weighted by $\lambda$ leads $\hat\z_0$ towards the fixed point. Unfortunately, due to parametrization and optimization errors in the VAE used for LDMs, we find that successive application of $\Sc(\cdot) := \Ec(\Dc(\cdot))$ does not lead to a meaningful fixed point. Rather, the images {\em always} diverge even if it started from a clean, natural in-distribution image. This is illustrated in Fig.~\ref{fig:fixed_point}, where we take 256 images from ImageNet validation set and measure the Euclidean distance $\|\x_{i+1} - \x_i\|_2^2$ for 25 iterations, where $\x_{i+1} = \Sc(\x_i)$. Due to this limitation in the pre-trained autoencoder, we find that the regularization in \eqref{eq:psld_approx} does not lead to consistent improvements in quality, or eliminate the atifacts that arise during the sampling process.

We take a different approach, simply constraining the clean latents to stay on the {\em range space} of the encoder $\Ec$ to minimize the train-test time discrepancy. This is natural as the training of LDMs are done with latents that are in the range space of $\Ec$. Moreover, mapping towards a lower-dimensional manifold typically results in removal of redundancies, which leads to the removal of artifacts. While the input to the encoder can be any $\x \in \Rd^n$, we constrain it to be 1) consistent with the measurement $\y$, and 2) close to the initial latent $\hat\z_0$. This leads to the following proximal optimization problem~\citep{parikh2014proximal},
\begin{align}
\label{eq:proximal}
    \bm{\Gamma}(\Dc(\hat\z_0)) := \x^* = \argmin_\x \frac{1}{2}\|\y - \Ab\x\|_2^2 + \frac{\lambda}{2}\|\x - \Dc(\hat\z_0)\|_2^2.
\end{align}
Subsequently, the mapping to the range space can be simply done through $\z^* = \Ec(\x^*)$. Notice that after the computation of $\Dc(\hat\z_0)$, \eqref{eq:proximal} does not involve any forward/backward pass through the neural network, and hence can be done with negligible computation cost using e.g. conjugate gradient (CG). In practice, we choose to apply our projection $\Ec(\bm{\Gamma}(\Dc(\hat\z_0))$ every few iteration to correct for the artifacts that arise during sampling, to control dramatic changes in sampling, and to save computation.

Nevertheless, solving \eqref{eq:proximal} requires access to $\Ab^\top$, which is often non-trivial to define. 
For instance, even for the widely-explored deblurring, correctly defining $\Ab^\top$ is tricky when the degradation model is not assumed to be a circular convolution.
Contrarily, in our \texttt{jax} implementation, $\Ab^\top$ can be implicitly defined through \texttt{jax.vjp}. Hence, the only requirement to be able to apply our method is the differentiability of the forward operator $\Ab$, similar to~\cite{chung2023diffusion}. For further discussion, see Appendix~\ref{sec:implementation_in_jax}.

\paragraph{Targetting arbitrary resolution}

Despite its fully convolutional nature, as SD was trained with 64$\times$64 latents ($\leftrightarrow 512\times 512$ images), the performance degrades when we aim to deal with larger dimensions, again due to train-test time discrepancy. Several works aimed to mitigate this issue by processing the latents with strided patches~\citep{bar2023multidiffusion,jimenez2023mixture,wang2023exploiting} that increases the computational burden by roughly $\mathcal{O}(n^2)$. In contrast, we show that our projection approach, used {\em without} any patch processing, can outperform previous methods that rely on patches, resulting in significantly faster inference speed.

\section{Experiments}

\begin{table}[!thb]
\centering
\setlength{\tabcolsep}{2pt}
\resizebox{1.0\textwidth}{!}{%
\begin{tabular}{lcccccccccccc}
\toprule
{} & \multicolumn{3}{c}{\textbf{SR ($\times 8$)}} & \multicolumn{3}{c}{\textbf{Deblur (motion)}} & \multicolumn{3}{c}{\textbf{Deblur (gauss)}} & \multicolumn{3}{c}{\textbf{Inpaint}}
\\
\cmidrule(lr){2-4}
\cmidrule(lr){5-7}
\cmidrule(lr){8-10}
\cmidrule(lr){11-13}
{\textbf{Method}} & {FID$\downarrow$} & {LPIPS$\downarrow$} & {PSNR$\uparrow$} & {FID$\downarrow$} & {LPIPS$\downarrow$} & {PSNR$\uparrow$} & {FID$\downarrow$} & {LPIPS$\downarrow$} & {PSNR$\uparrow$} & {FID$\downarrow$} & {LPIPS$\downarrow$} & {PSNR$\uparrow$}\\
\midrule
\rowcolor{BrickRed!10}
P2L~(ours) & \textbf{31.23} & \textbf{0.290} & \underline{28.55} & \underline{28.34} & \textbf{0.302} & \textbf{27.23} & \textbf{30.62} & \textbf{0.299} & 26.97 & \textbf{26.27} & \textbf{0.168} & \underline{25.29}\\
\cmidrule(l){1-13}
LDPS & 36.81 & \underline{0.292} & \textbf{28.78} & 58.66 & 0.382 & 26.19 & 45.89 & 0.334 & 27.82 & 46.10 & 0.311 & 23.07\\
GML-DPS~\citep{rout2023solving} & 41.65 & 0.318 & 28.50 & 47.96 & 0.352 & \underline{27.16} & 42.60 & 0.320 & \textbf{28.49} & 36.31 & \underline{0.208} & 23.10\\
PSLD~\citep{rout2023solving} & 36.93 & 0.335 & 26.62 & 47.71 & 0.348 & 27.05 & 41.04 & 0.320 & \underline{28.47} & 35.01 & 0.207 & 23.10\\
LDIR~\citep{he2023iterative} & \underline{36.04} & 0.345 & 25.79 & \textbf{24.40} & 0.376 & 24.40 & \underline{35.61} & 0.341 & 25.75 & 37.23 & 0.250 & \textbf{25.47}\\
\cmidrule(l){1-13}
\rowcolor{gray!10}
DDS~\citep{chung2023fast} & 262.0 & 1.278 & 13.01 & 88.70 & 1.014 & 14.68 & 74.02 & 0.932 & 17.03 & 113.6 & 0.421 & 17.92\\
\rowcolor{gray!10}
DPS~\citep{chung2023diffusion} & 47.65 & 0.340 & 21.81 & 65.91 & 0.601 & 21.11 & 100.2 & 0.983 & 15.71 & 137.7 & 0.692 & 15.35\\
\rowcolor{gray!10}
DiffPIR~\citep{zhu2023denoising} & 141.1 & 1.266 & 13.80 & 72.02 & 0.664 & 21.03 & 69.15 & 0.751 & 22.27 & \underline{33.92} & 0.238 & 24.91\\
\bottomrule
\end{tabular}
}
\caption{
Quantitative evaluation (PSNR, LPIPS, FID) of inverse problem solving on FFHQ 512$\times$512-1k validation dataset. \textbf{Bold}: best, \underline{underline}: second best. Methods that are not LDM-based are shaded in gray.
}
\label{tab:results_ffhq}
\end{table}

\subsection{Experimental setting}

\paragraph{Datasets, Models}
We consider two different well-established datasets: 1) FFHQ 512$\times$512~\citep{karras2019style}, and 2) ImageNet 512$\times$512~\citep{deng2009imagenet}. For the former, we use the first 1000 images for testing, similar to \cite{chung2023diffusion}. For the latter, we choose 1k images out of 10k test images provided in~\cite{saharia2022palette} by interleaved sampling, i.e. using images of index 0, 10, 20, etc. after ordering by name.
For the latent diffusion model, we choose SD v1.4 pre-trained on the LAION dataset for all the experiments, including the baseline comparison methods based on LDM. As there is no publicly available image diffusion model that is trained on an identical dataset, we choose ADM~\citep{dhariwal2021diffusion} trained on ImageNet 512$\times$512 data as the universal prior when implementing baseline image-domain DIS. 
Note that this discrepancy may lead to an unfair advantage in the performance for evaulation on ImageNet, and an unfair disadvantage in the performance when evaluating on FFHQ.
All experiments were done on NVIDIA A100 40GB GPUs.

\paragraph{Inverse Problems}
We test our method on the following degradations: 1) Super-resolution from $\times 8$ averagepooling, 2) Inpainting from 10-20\% free-form masking as used in~\cite{saharia2022palette}, 3) Gaussian deblurring from an image convolved with a 61$\times$61 size Gaussian kernel with $\sigma = 3.0$, 4) Motion deblurring from an image convolved with a 61$\times$61 motion kernel that is randomly sampled with intensity 0.5\footnote{\url{https://github.com/LeviBorodenko/motionblur}}, following~\cite{chung2023diffusion}. For all degradations, we include mild additive white Gaussian noise with $\sigma_y = 0.01$.

\paragraph{Evaluation}
As the main objective of this study is to improve the performance of LDIS, we mainly focus our evaluation on the comparison against the current SOTA LDIS: we compare against LDPS, GML-DPS~\citep{rout2023solving}, PSLD~\citep{rout2023solving}, and LDIR~\citep{he2023iterative}. 
Notably, all LDIS including the proposed P2L use 1000 NFE DDIM sampling with $\eta = 0.0$\footnote{The parameter $\eta$ indicates the stochasticity of the sampler. $\eta = 0.0$ leads to deterministic sampling.}, where the value of $\eta$ in \eqref{eq:dds} was found through grid search.
We additionally compare against SOTA image-domain DIS which can cope with noisy inverse problems without computing the SVD of the forward operator: DPS~\citep{chung2023diffusion}, Diff-PIR~\citep{zhu2023denoising}, and DDS~\citep{chung2023fast}. 
For DPS, we use 1000 NFE DDIM sampling. For Diff-PIR and DDS, we use 100 NFE DDIM sampling. We choose the optimal $\eta$ values for these algorithms through grid-search. Details about the comparison methods can be found in Appendix~\ref{sec:comp_methods}.
We perform quantitative evaluation with standard metrics: PSNR, FID, and LPIPS.

\begin{figure}[!t]
    \centering
    \includegraphics[width=1.0\textwidth]{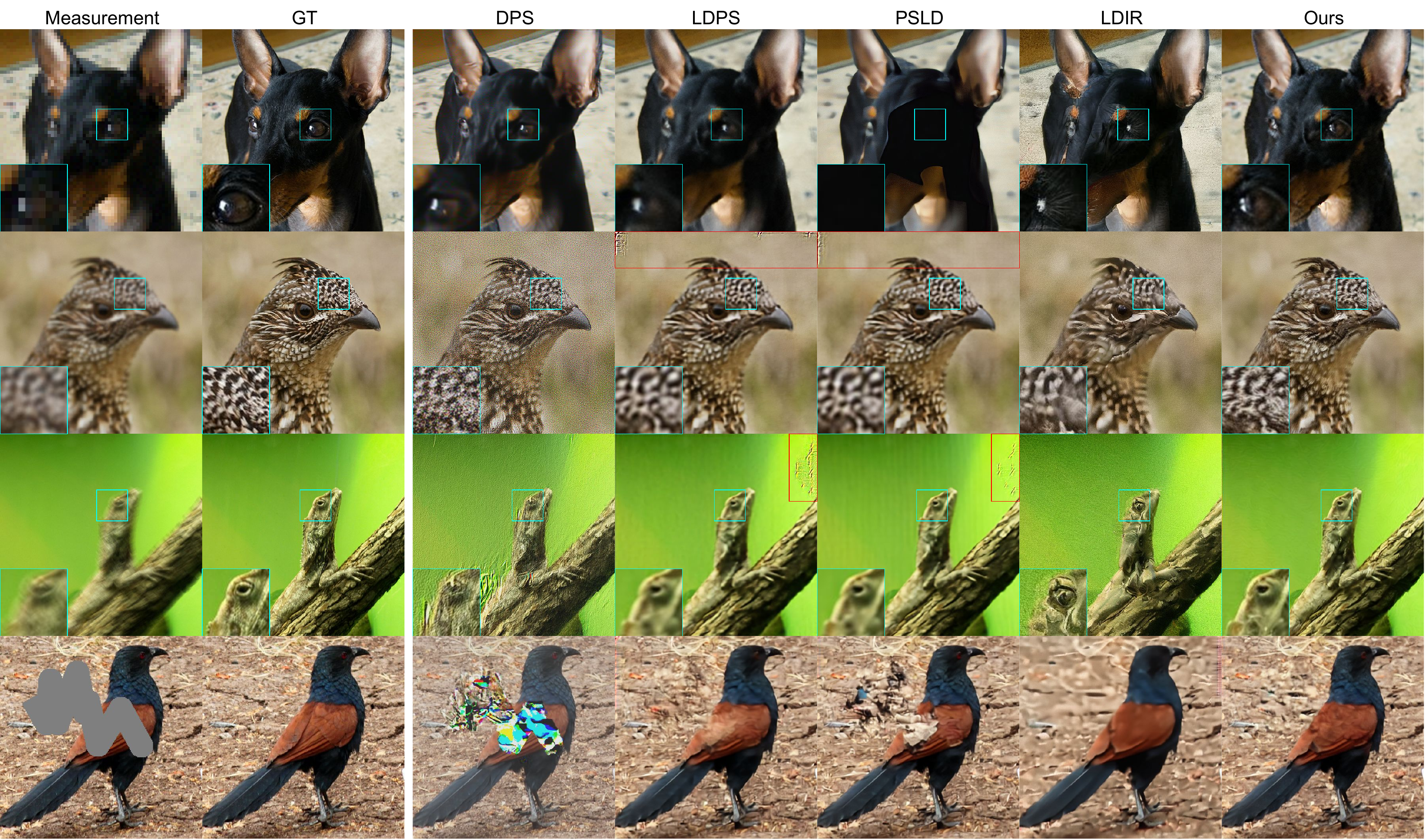}
    \caption{Inverse problem solving results on ImageNet $512\times 512$ test set. Row 1: SR$\times 8$, Row 2: gaussian deblurring, Row 3: motion deblurring, row 4: inpainting.}
    \vspace{-0.5cm}
    \label{fig:results_imagenet}  
\end{figure}

\begin{table}[!thb]
\setlength{\tabcolsep}{2pt}
\centering
\resizebox{1.0\textwidth}{!}{%
\begin{tabular}{lcccccccccccc}
\toprule
{} & \multicolumn{3}{c}{\textbf{SR ($\times 8$)}} & \multicolumn{3}{c}{\textbf{Deblur (motion)}} & \multicolumn{3}{c}{\textbf{Deblur (gauss)}} & \multicolumn{3}{c}{\textbf{Inpaint}}
\\
\cmidrule(lr){2-4}
\cmidrule(lr){5-7}
\cmidrule(lr){8-10}
\cmidrule(lr){11-13}
{\textbf{Method}} & {FID$\downarrow$} & {LPIPS$\downarrow$} & {PSNR$\uparrow$} & {FID$\downarrow$} & {LPIPS$\downarrow$} & {PSNR$\uparrow$} & {FID$\downarrow$} & {LPIPS$\downarrow$} & {PSNR$\uparrow$} & {FID$\downarrow$} & {LPIPS$\downarrow$} & {PSNR$\uparrow$}\\
\midrule
\rowcolor{BrickRed!10}
P2L~(ours) & \textbf{51.81} & \textbf{0.386} & \textbf{23.38} & \textbf{54.11} & \textbf{0.360} & \textbf{24.79} & \textbf{39.10} & \textbf{0.325} & 25.11 & \textbf{32.82} & \textbf{0.229} & \underline{21.99}\\
\cmidrule(l){1-13}
LDPS & 61.09 & 0.475 & \underline{23.21} & 71.12 & 0.441 & 23.32 & 48.17 & 0.392 & 24.91 & 46.72 & 0.332 & 21.54\\
GML-DPS~\citep{rout2023solving} & 60.36 & \underline{0.456} & \underline{23.21} & 59.08 & 0.403 & 24.35 & \underline{45.33} & 0.377 & \textbf{25.44} & 47.30 & 0.294 & 21.12\\
PSLD~\citep{rout2023solving} & 60.81 & 0.471 & 23.17 & 59.63 & 0.398 & 24.21 & 45.44 & \underline{0.376} & \underline{25.42} & \underline{40.57} & \underline{0.251} & 20.92\\
LDIR~\citep{he2023iterative} & 63.46 & 0.480 & 22.23 & 88.51 & 0.475 & 21.37 & 72.10 & 0.506 & 22.45 & 50.65 & 0.313 & \textbf{23.28}\\
\cmidrule(l){1-13}
\rowcolor{gray!10}
DDS~\citep{chung2023fast} & 203.2 & 1.213 & 12.72 & 84.67 & 0.925 & 14.52 & 70.51 & 0.835 & 16.58 & 60.18 & 0.354 & 17.03\\
\rowcolor{gray!10}
DPS~\citep{chung2023diffusion} & \underline{54.61} & 0.544 & 20.70 & 71.99 & 0.599 & 19.62 & 98.33 & 0.910 & 15.05 & 71.70 & 0.360 & 15.15\\
\rowcolor{gray!10}
DiffPIR~\citep{zhu2023denoising} & 488.3 & 1.182 & 13.44 & 87.04 & 0.622 & 19.32 & 79.31 & 0.755 & 20.55 & 45.97 & 0.300 & 20.11\\
\bottomrule
\end{tabular}
}
\caption{
Quantitative evaluation (PSNR, LPIPS, FID) of inverse problem solving on ImageNet 512$\times$512-1k validation dataset. \textbf{Bold}: best, \underline{underline}: second best. Methods that are not LDM-based are shaded in gray.
}
\label{tab:results_imagenet}
\end{table}

\subsection{Main results}

\paragraph{Comparison against baseline}
In all of the inverse problems that we consider in the paper, our method outperforms all the baselines by quite a large margin in terms of perceptual quality, measured by FID and LPIPS, while keeping the distortion at a comparable level against the current state-of-the-art methods. Especially, we see about 10 FID decrease in deblurring and inpainting tasks compared to the runner up in both FFHQ and ImageNet dataset (See Tables~\ref{tab:results_ffhq},\ref{tab:results_imagenet}). The superiority can also be clearly seen in Fig.~\ref{fig:results_imagenet}, where P2L achieves stable, high-quality reconstruction throughout all tasks. Results from both LDPS and PSLD often contain local grid-like artifacts (Red boxes in Figures) and are blurry. With P2L, the restored images are sharpened while the artifacts are effectively removed. LDIR are less prone to artifacts owing to the smoothed history gradient updates, but often results in unrealistic textures and deviations from the measurement, which is also reflected in having the lowest PSNR among the LDIS-class methods. In contrast, P2L is free from such drawbacks even when leveraging Adam-like gradient update steps.

One rather surprising finding is the heavy downgrade in the performance for DIS methods. Even on in-distribution ImageNet test data, methods such as DPS and DiffPIR become very unstable. This can be attributed to the generative prior being poor: directly training diffusion models on high-resolution images often result in poor performance\footnote{Consequently, for $\geq 512\times 512$ resolution, either using latent diffusion or using cascaded models~\citep{saharia2022photorealistic} are popular.}. This observation again points to the importance of developing methods that can leverage foundation models when aiming for general domain higher-resolution data. See Appendix~\ref{sec:further_experimental_results} for further results.
As a final note, we believe that the compromise in PSNR is related to the imperfectness of the VAE used in SD v1.4\footnote{Auto-encoding 1000 ground-truth test images result in the following metrics: FFHQ (PSNR): 29.66 $\pm$ 2.29, ImageNet (PSNR): 27.12 $\pm$ 4.38.}, and we expect such degradation to be mitigated when switching to better, larger autoencoders such as SDXL~\citep{podell2023sdxl}. 

\paragraph{Design components}

\begin{table}[!t]
    \centering
    \setlength{\tabcolsep}{2pt}
    \begin{minipage}{0.77\textwidth}
        \centering
        \resizebox{\textwidth}{!}{%
        \begin{tabular}{cccllllllll}
            \toprule
            {} & {} & {} & \multicolumn{4}{c}{\textbf{FFHQ}} & \multicolumn{4}{c}{\textbf{ImageNet}} \\
            \cmidrule(lr){4-7}
            \cmidrule(lr){8-11}
            \multicolumn{3}{c}{Design components} & \multicolumn{2}{c}{SR$\times 8$} & \multicolumn{2}{c}{Inpaint ($p=0.8$)} & \multicolumn{2}{c}{SR$\times 8$} & \multicolumn{2}{c}{Inpaint ($p=0.8$)} \\
            \cmidrule(lr){1-3}
            \cmidrule(lr){4-5}
            \cmidrule(lr){6-7}
            \cmidrule(lr){8-9}
            \cmidrule(lr){10-11}
            {Projection} & {$\bm{\Gamma}$} & {Prompt tuning} & {FID$\downarrow$} & {PSNR$\uparrow$} & {FID$\downarrow$} & {PSNR$\uparrow$} & {FID$\downarrow$} & {PSNR$\uparrow$} & {FID$\downarrow$} & {PSNR$\uparrow$}\\
            \midrule
            \xmark & \xmark & \xmark & 61.16 & 26.49 & 52.34 & \textbf{29.78} & 78.68 & 23.49 & 70.87 & 26.20 \\
            \xmark & \xmark & \cmark & 58.73 & \textbf{26.68} & 51.40 & 29.69 & 76.40 & \textbf{23.52} & 67.06 & 26.32 \\
            \cmark & \xmark & \xmark & 55.91 & 26.37 & 48.71 & 29.68 & 74.22 & 23.16 & 66.92 & 26.08 \\
            \cmark & \cmark & \xmark & \underline{55.68} & 26.43 & \underline{47.76} & \underline{29.70} & \underline{74.01} & 23.32 & 65.45 & 26.29 \\
            \rowcolor{BrickRed!10}
            \cmark & \cmark & \cmark & \textbf{52.96} & \underline{26.64} & \textbf{46.92} & 29.63 & \textbf{70.08} & 23.48 & \textbf{59.26} & 26.12 \\
            \bottomrule
        \end{tabular}
        }
        \caption{Ablation studies on the design components}
        \vspace{-0.5cm}
        \label{tab:ablation_design_component}
    \end{minipage}\hfill
    \begin{minipage}{0.225\textwidth}
        \centering
        \resizebox{\textwidth}{!}{%
        \begin{tabular}{ccll}
            \toprule
            $\sigma_y$ & $\bm{\Gamma}$ & PSNR & FID \\
            \midrule
            \multirow{2}{*}{0.0} & glue & 26.51 & 54.69 \\
            & \cellcolor{BrickRed!10}{Ours} & \cellcolor{BrickRed!10}{\textbf{26.80}} & \cellcolor{BrickRed!10}{\textbf{54.58}} \\
            \midrule
            \multirow{2}{*}{0.01} & glue & 26.39 & 56.47 \\
            & \cellcolor{BrickRed!10}{Ours} & \cellcolor{BrickRed!10}{\textbf{26.43}} & \cellcolor{BrickRed!10}{\textbf{55.68}} \\
            \midrule
            \multirow{2}{*}{0.05} & glue & 23.86 & 68.99 \\
            & \cellcolor{BrickRed!10}{Ours} & \cellcolor{BrickRed!10}{\textbf{24.92}} & \cellcolor{BrickRed!10}{\textbf{65.90}} \\
            \bottomrule
        \end{tabular}
        }
        \caption{Choice of $\bm{\Gamma}$}
        \vspace{-0.5cm}
        \label{tab:choice_of_gamma}
    \end{minipage}
\end{table}

In Table~\ref{tab:ablation_design_component}, we perform an ablation study on the design components of the proposed method. From the table, we confirm that prompt tuning, projection to the range space of the encoder, and performing proximal update step (denoted as $\bm{\Gamma}$) before the projection all contributes to the gain in the performance. It is important that these gains are synergistic, and one component does not hamper the other. 
In the Appendix Tab.~\ref{tab:pt_robustness}, we further show that our prompt-tuning approach is robust to the variation in the hyper-parameters (learning rate, number of iterations). Specifically, among the 9 configurations that we try, only the one with 5 iterations, lr=0.001 is inferior to not using prompt tuning. 
In Fig.~\ref{fig:progress}, we visualize the progress of $\Dc(\hat\z_0)$ through time $t$ starting from the same random seed, comparing LDPS, PSLD, and LDPS + projection (row 4 of Tab.~\ref{tab:hparam}). Here, we see that our proposed projection approach effectively suppresses the artifacts that arise during the reconstruction process, whereas PSLD introduces additional artifacts.

\paragraph{Choice of $\bm{\Gamma}$}
When projecting to the range space of $\Ec$, we choose to use the proximal optimization strategy in \eqref{eq:proximal}. Instead, one could resort to projection to the measurement subspace (``gluing'' of~\cite{rout2023solving}) by using $\bm{\Gamma}(\hat\x_0) = \Ab^\top\y + (\Ib - \Ab^\top\Ab)\hat\x_0$. In Table~\ref{tab:choice_of_gamma}, we compare our choice of $\bm{\Gamma}$ against the gluing on various noise levels on FFHQ SR$\times 8$. We see that for all noise levels, the proximal steps consistently outperform the gluing, even when $\bm{\Gamma}$ is applied every $\gamma = 4$ steps of reverse diffusion. Furthermore, due to the noise-amplifying nature of projection, the differences become more pronounced as we increase the noise level.

\begin{figure}[!t]
    \centering
    \begin{tikzpicture}
        \node[anchor=south west,inner sep=0] (image) at (0,0) {\includegraphics[width=1.0\textwidth]{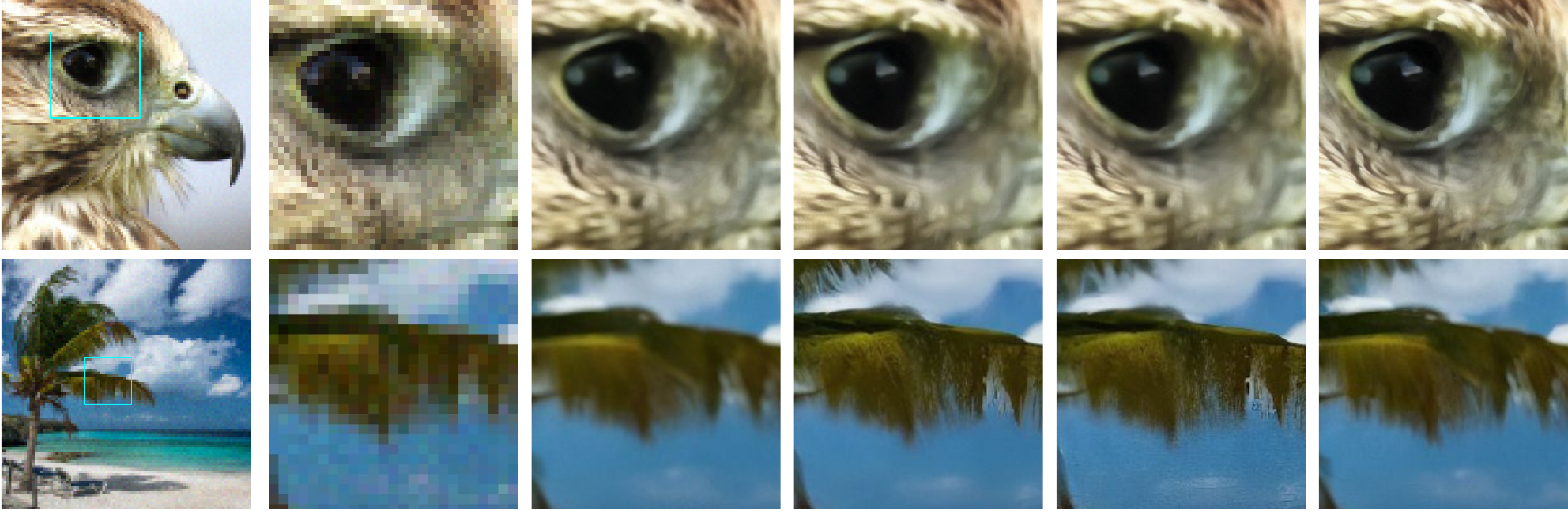}};
        \begin{scope}[x={(image.south east)},y={(image.north west)}]
            \node at (0.42,1.05) {{\tiny Vanilla [1]}};
            \node at (0.59,1.05) {{\tiny\cite{bar2023multidiffusion} [4]}};
            \node at (0.76,1.05) {{\tiny\cite{jimenez2023mixture} [4]}};
            \node at (0.92,1.05) {{\tiny Proposed [1]}};
        \end{scope}
    \end{tikzpicture}
    \caption{Results on $\times 8$ SR on DIV2K validation set of 768$\times$768 resolution. [Diffusion NFE per denoising step]. Vanilla and proposed process the latent as a whole.}
    \vspace{-0.5cm}
    \label{fig:highres_exp}
\end{figure}

\paragraph{High-resolution restoration}
In Fig.~\ref{fig:highres_exp}, we show the effectiveness of our projection method on arbitrary resolution image restoration by comparing our method to \cite{bar2023multidiffusion} and \cite{jimenez2023mixture}, as well as the case where the larger latents are processed as a whole without patching (denoted as vanilla). Here, we see that the proposed method provides the best result, even when we use 1 NFE per denoising step, in contrast to requiring 4 NFE per denoising in the comparison methods. Further details and discussion is provided in Appendix~\ref{sec:highres}.

\section{Conclusion}

We proposed P2L, a latent diffusion model-based inverse problem solver that introduces two new strategies. First, a prompt tuning method to optimize the continuous input text embedding used for diffusion models was developed. We observed that our strategy can boost the performance by a good margin compared to the usage of null text embedding that prior works employ. Second, a projection approach to keep the latents in the range space of the encoder during the reverse diffusion process was proposed. Our approach effectively mitigated the artifacts that often arise during inverse problem solving, while also sharpening the final output. P2L outperforms previous diffusion model-based inverse problem solvers that operate on the latent and the image domain.

\paragraph{Limitations}
While prompt tuning enhances the performance, it also incurs additional computational complexity as additional forward/backward passes through the latent diffusion model and the decoder is necessary. Consequently, the method will need future investigations when aiming for time-critical applications. 
As we optimize the continuous text embeddings rather than the discrete text directly, it is hard to decipher what the text embedding after the optimization has converged to. This is a limitation of the text embedder used for SD, as CLIP does not utilize a decoder. We could instead opt for the use of Imagen~\citep{saharia2022photorealistic}, where T-5 with an encoder-decoder architecture is used, where one could easily check the learned text from our prompt-tuning scheme.
Moreover, we did not consider the usage of CFG, which would enable flexible control on the degree of sharpening. Extending the prompt tuning idea to jointly optimize the embedding of the conditional and the unconditional model may be an interesting direction of future research.

\paragraph{Ethics statement}
While our method can lead to advancements in areas such as computational imaging, medical imaging, and other fields where inverse problems are prevalent, we also recognize the potential for misuse in areas like deepfake generation or unauthorized data reconstruction, naturally leading from the use of generative models. The potential bias within the training dataset of the diffusion model may be potentially amplified with the usage of our method.
We have taken care to ensure that our experiments adhere to ethical guidelines, using publicly available datasets or those for which we have obtained explicit permissions. We urge the community to adopt responsible practices when applying our findings and to consider the broader societal implications of the technology.

\paragraph{Reproducibility statement}
In order to facilitate reproducibility, We detail our implementation in the form of Algorithms (Alg.~\ref{alg:fn:optimize_emb},\ref{alg:p2l},\ref{alg:p2l_adam}), and pseudo-code (Fig.~\ref{fig:jax_AT}). The specific hyper-parameters chosen for the method is detailed in Appendix~\ref{sec:imp_details}.

\bibliography{iclr2024_conference}
\bibliographystyle{iclr2024_conference}

\clearpage

\appendix
\section{Proof-of-concept experiment}
\label{sec:proof_of_concept_exp}

\begin{figure}[!t]
    \centering
    \includegraphics[width=1.0\textwidth]{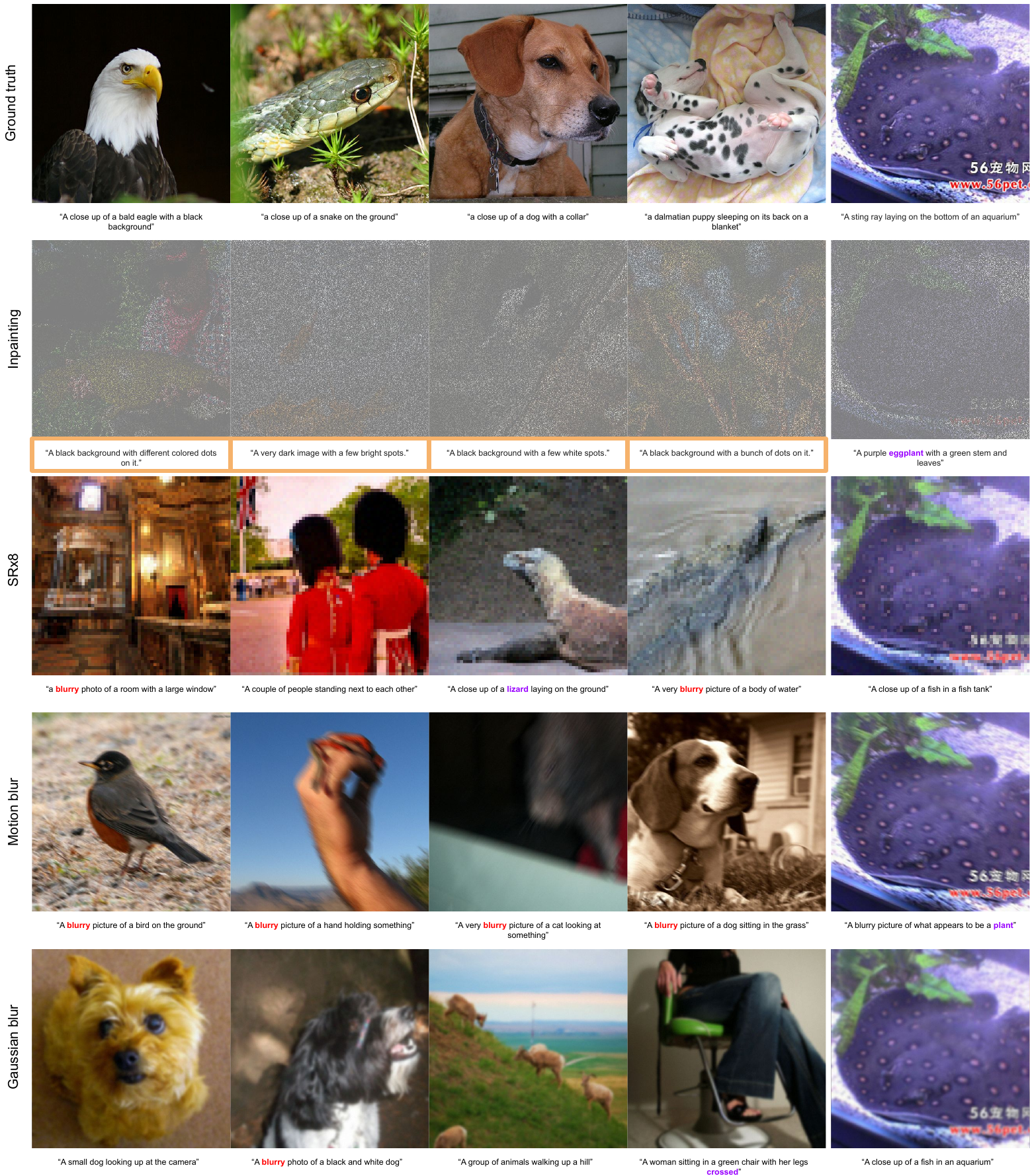}
    \caption{Captions generated by PALI~\citep{chen2022pali} from ground-truth ImageNet 512$\times$512 clean images, and the degraded images. The rightmost column contain images that are from the same ground truth. Captions in in orange box completely fail to describe the underlying image. {\color{xkcdPurple}{Purple}} captions wrongly identify the image. Captions generated from degraded measurements often contain negative words such as \color{red}{blurry}.}
    \label{fig:pali_oracle_imagenet}  
\end{figure}

For the caption generation with PALI, we simply take the captions with the highest score. Examples of the captions generated from PALI are presented in Fig.~\ref{fig:pali_oracle_imagenet}. In our initial experiments, we found that using PALI captions directly did not directly lead to an improvement in the performance, as it only describes the {\em content} of the image, and says nothing about the {\em quality} of the image. Therefore, we use the following text prompts for the oracle ``\code{A high quality photo of a} \{PALI\_prompt\}'', similar to the general text prompts.

For both inverse problems (SR$\times 8$, inpainting with $p = 0.8$), we use the LDPS algorithm with 1000 NFE and $\eta = 0.0$. We apply prompt tuning algorithm per denoising step as indicated in Algorithm~\ref{alg:fn:optimize_emb}, with $K = 5$ and learning rate of $1e-4$. When optimizing for the text embedding, we initialize it with the embedding vector from the token ``\code{A high quality photo of a face}'' for FFHQ, and ``\code{A high quality photo}'' for ImageNet in the case of inpainting. Note that for the latter, we did not find much performance difference when initializing from the null text prompt, or even initializing it with ``\code{A high quality photo of a dog}''. For $\times 8$ SR, we initialize the text embeddings from PALI captions generated from $\y$, as we empirically observe that PALI captions from $\y$ still have a relatively good coarse description about the given image.

\section{Implementation details}
\label{sec:imp_details}

\begin{table}[t]
\centering
\resizebox{1.0\textwidth}{!}{%
\begin{tabular}{lllllllll}
\toprule
{} & \multicolumn{4}{c}{\textbf{FFHQ}} & \multicolumn{4}{c}{\textbf{ImageNet}}\\
\cmidrule{2-5}
\cmidrule{6-9}
problem & Deblur (motion) & Deblur (gauss) & SR$\times 8$ & inpaint & Deblur (motion) & Deblur (gauss) & SR$\times 8$ & inpaint \\
\midrule
Gradient type & Adam & Adam & GD & Adam & Adam & GD & GD & GD \\
$\rho_t$ & 0.05 & 0.05 & 1.0 & 0.05 & 0.1 & $\bar\alpha_t$ & $15 \bar\alpha_t$ & 0.5 \\
\rowcolor{mylightblue}
$\gamma$ & 5 & 4 & 4 & 3 & 5 & 4 & 4 & 3 \\
\rowcolor{mylightblue}
$\lambda$ & 1.0 & 1.0 & 1.0 & 0.1 & 1.0 & 1.0 & 1.0 & 0.1 \\
\rowcolor{BrickRed!10}
$K$ & 3 & 5 & 5 & 1 & 3 & 3 & 3 & 1 \\
\rowcolor{BrickRed!10}
learning rate & $5e-5$ & $1e-4$ & $1e-4$ & $1e-4$ & $1e-5$ & $1e-4$ & $1e-5$ & $1e-4$ \\
\bottomrule
\end{tabular}
}
\caption{
Hyper-parameter choice for the proposed method. White shade: hyper-parameters related to gradient updates, blue shade: hyper-parameters related to projecting onto the range space of $\Ec$, red shade: hyper-parameters related to prompt tuning.
}
\label{tab:hparam}
\end{table}

\subsection{Step 1: $\Cc$ update prompt tuning}

Since we do not have the ground truth clean image to optimize the conditional embedding over, we use the following optimization strategy
\begin{align}
\label{eq:pt}
    \Cc^* = \argmin_{\Cc} \|\Ab\Dc\left(\Ed[\z_0|\z_t, \y]\right) - \y\|_2^2,
\end{align}
where \eqref{eq:pt} is performed for every timestep $t$ during the inference stage. Here, we approximate the conditional posterior mean as
\begin{align}
    \Ed[\z_0|\z_t,\y] &= \frac{1}{\sqrt{\bar\alpha_t}} \z_t + \frac{1 - \bar\alpha_t}{\sqrt{\bar\alpha_t}}
    \left( \nabla_{\z_t} \log p(\z_t) + \nabla_{\z_t} \log p(\y|\z_t) \right) \\
    &\simeq \Ed[\z_0|\z_t] + \frac{1 - \bar\alpha_t}{\sqrt{\bar\alpha_t}} \nabla_{\hat\z_{0|t}} \log p(\y|\hat\z_{0|t}),
\end{align}
which is the result of the approximations proposed in~\citep{ravula2023optimizing,barbano2023steerable}. In practice, we choose a static step size $\rho = 1.0$ with the gradient of the norm, which was shown to be effective in~\citep{chung2023diffusion}. The resulting prompt tuning algorithm is summarized in Algorithm~\ref{alg:fn:optimize_emb}. Notice that we update our embeddings to improve the fidelity \eqref{eq:pt}. However, in practice, this also leads to higher quality images in terms of perception.
For optimizing \eqref{eq:pt}, we use Adam with the learning rate and the number of iterations as denoted in Table~\ref{tab:hparam} for every $t$.

\begin{algorithm}[H]
    \small
    \caption{P2L: Adam}
    \label{alg:p2l_adam}
    \begin{algorithmic}[1]
    \Require $\epsilonb_{\thetab^{*}}, \z_T, \y, \Cc, T, K, \gamma, \beta_1, \beta_2, \varepsilon, \bm{\Gamma}$
    \State $\bm{m}_T \gets$ \code{np.zeros\_like}$({\zb_T})$
    \State $\bm{v}_T \gets$ \code{np.zeros\_like}$({\zb_T})$
    \For{$t=T$ {\bfseries to} $1$}
        \State $\Cc_t^{*} \gets \color{xkcdWine}{\textsc{OptimizeEmb}}(\z_t, \y, \Cc_t^{0})$
        \State $\hat\epsilonb_{t} \gets \epsilonb_{\thetab^{*}}(\z_{t}, \Cc_t^{*})$
        \State $\hat\z_{0|t} \gets (\z_t - \sqrt{1 - \bar\alpha_t} \hat{\epsilonb}_{t}){/\sqrt{\bar\alpha_t}}$
        \If {$(t \mod \gamma) = 0$}
            \State $\color{xkcdOrange}{\hat\z'_{0|t} \gets \Ec\left({\bm\Gamma}\left(\Dc(\hat\z_{0|t})\right)\right)}$
        \EndIf
        \State $\z'_{t-1} \gets \sqrt{\bar\alpha_{t-1}}\hat\z'_{0|t} + \sqrt{1 - \bar\alpha_{t-1}}\hat\epsilonb_t$
        \State $\gb \gets \nabla_{\z_t}\|\Ac\Dc(\hat\z_{0|t}) - \y\|$
        \State $\hat{\bm{m}}_{t-1} \gets \left(\beta_1 \bm{m}_t + (1 - \beta_1)\gb\right) / (1 - \beta_1)$
        \State $\hat{\bm{v}}_{t-1} \gets \left(\beta_2 \bm{v}_t + (1 - \beta_2)(\gb \circ \gb)\right) / (1 - \beta_2)$
        \State $\z_{t-1} \gets \z'_{t-1} - \rho_{t} \frac{\hat{\bm{m}}_{t-1}}{\sqrt{\hat{\bm{v}}_{t-1}} + \varepsilon}$
        \State $\Cc_{t-1}^{(0)} \gets \Cc_t^*$
        \EndFor
    \State {\bfseries return} $\x_0 \gets \Dc(\z_0)$
    \end{algorithmic}
\end{algorithm}

\subsection{Step 2: $\z_t$ update}

In Table~\ref{tab:hparam}, there are two gradient types: GD and Adam. For GD, we use standard gradient descent steps as presented in Algorithm~\ref{alg:p2l}. For Adam, using the same prompt tuning Algorithm~\ref{alg:fn:optimize_emb}, we adopt a history gradient update scheme as proposed in \cite{he2023iterative} to arrive at Algorithm~\ref{alg:p2l_adam}. Note that the hyper-parameters of the Adam update were fixed to be $\beta_1 = 0.9, \beta_2 = 0.999, \varepsilon=1e-8$, which is the default setting. We only search for the optimal step size $\rho_t$ via grid search, which is set to 0.1 for motion deblurring in ImageNet, and 0.05 otherwise.

\subsection{Comparison methods}
\label{sec:comp_methods}

\paragraph{LDPS}

LDPS can be considered a straightforward extension image domain DPS~\citep{chung2023diffusion}. The three works that we review in this section~\citep{he2023iterative,rout2023solving,song2023solving} all consider LDPS as a baseline. In LDPS, we have the following update scheme

\begin{align}
    \z_{t-1} &= {\rm DDIM}(\z_t) - \rho \nabla_{\z_t} \|\y - \Ab\Dc(\hat\z_0)\|_2,
\end{align}
where $\rho$ is the step size, and ${\rm DDIM}(\cdot)$ denotes a single step of DDIM sampling. We use a static step size of $\rho = 1$, widely adopted in literature.

\paragraph{LDIR~\citep{he2023iterative}}

Using Adam-like history gradient update scheme, a single iteration of the algorithm can be summarized as follows
\begin{align}
    \g_t &= \nabla_{\z_t} \|\y - \Ab\Dc(\hat\z_0)\| \\
    \hat\m_t &= (\beta_1 \m_{t-1} + (1 - \beta_1) \g_t) / (1 - \beta_1) \\
    \hat\vv_t &= (\beta_2 \vv_{t-1} + (1 - \beta_2) (\g_t \circ \g_t)) / (1 - \beta_2) \\
    \z_{t-1} &= {\rm DDIM}(\z_t) - \rho \frac{\hat\m_t}{\sqrt{\hat\vv_t} + \varepsilon},
\end{align}
where $\circ$ denotes element-wise product, and $\beta_1, \beta_2, \varepsilon$ are the hyperparameters of the sampling scheme. As LDIR uses a momentum-based update scheme, we have smoother gradient transitions. We fix $\beta_1=0.9, \beta_2=0.999, \varepsilon=1e-8$ to be identical to when using the proposed method. The step size $\rho$ is chosen to be the optimal value found through grid search: 0.1 for ImageNet motion deblurring, and 0.05 otherwise.

\paragraph{GML-DPS, PSLD~\citep{rout2023solving}}

GML-DPS attempts to regularize the predicted clean latent $\hat\z_0$ to be a fixed point after encoding and decoding. Formally, the update step reads
\begin{align}
\label{eq:gml-dps}
    \z_{t-1} &= {\rm DDIM}(\z_t) - \rho \nabla_{\z_t} \left( \|\y - \Ab\Dc(\hat\z_0)\|_2 + \gamma\|\hat\z_0 - \Ec(\Dc(\hat\z_0))\|_2 \right).
\end{align}
Further, PSLD applies an orthogonal projection onto the subspace of $\Ab$ in between decoding and encoding to enforce fidelity
\begin{align}
\label{eq:psld}
    \z_{t-1} &= {\rm DDIM}(\z_t) - \rho \nabla_{\z_t} \left( \|\y - \Ab\Dc(\hat\z_0)\|_2 + \gamma\|\hat\z_0 - \Ec(\Ab^\top\y + (\Ib - \Ab^\top\Ab)\Dc(\hat\z_0))\|_2 \right).
\end{align}
We use the static step size of $\rho = 1$, and choose $\gamma = 0.1$, as advised in~\cite{rout2023solving}.
GML-DPS and PSLD are closest to the proposed method in spirit, as these methods attempt to guide the latents to stay closer to the natural manifold by enforcing them to be a fixed point after autoencoding. The difference is that these approaches use gradient guidance while we try to explicitly project the latents into the the natural manifold.

\paragraph{DPS~\citep{chung2023diffusion}}
DPS is a DIS that utilizes the following update scheme\footnote{The original work only considered DDPM sampling. We consider DDIM as a generalization of DDPM as it can be retrieved with $\eta = 1.0$.}
\begin{align}
\label{eq:dps}
    \x_{t-1} &= {\rm DDIM}(\x_t) - \nabla_{\x_t} \left( \|\y - \Ab\hat\x_0\|_2 \right).
\end{align}
The optimal value of $\eta$ was found through grid search for each inverse problem: $\eta = 0.0$ for SR$\times 8$, and $\eta = 1.0$ for others.

\paragraph{DDS~\citep{chung2023fast}}
The following updates are used
\begin{align}
\label{eq:dds}
    \hat\x'_0 &= \argmin_\x \frac{1}{2} \|\y - \Ab\xb\|_2^2 + \frac{\gamma}{2} \|\x - \hat\x_0\|_2^2 \\
    \x_{t-1} &= \sqrt{\bar\alpha_{t-1}}\hat\x'_0 + \sqrt{1 - \bar\alpha_{t-1} - \eta^2 \tilde\beta_{t-1}^2}\hat\epsilonb_t + \eta \tilde\beta_{t-1}\epsilonb,
\end{align}
where \eqref{eq:dds} is solved through CG with 5 iterations, $\gamma = 1.0$. $\eta = 0.0$ is chosen for Gaussian deblurring, and $\eta = 1.0$ for the rest of the inverse problems.

\paragraph{DiffPIR~\citep{zhu2023denoising}}
Similar to DDS, the following updates are used
\begin{align}
\label{eq:diffpir}
    \hat\x'_0 &= \argmin_\x \frac{1}{2} \|\y - \Ab\xb\|_2^2 + \frac{\lambda\sigma^2\bar\alpha_t}{2(1 - \bar\alpha_t)} \|\x - \hat\x_0\|_2^2 \\
    \x_{t-1} &= \sqrt{\bar\alpha_{t-1}}\hat\x'_0 + \sqrt{1 - \bar\alpha_{t-1}}(\sqrt{1 - \zeta}\hat\epsilonb_t + \sqrt{\zeta}\epsilonb),
\end{align}
where $\sigma$ is the noise level of the measurement, and $\lambda,\zeta$ are hyper-parameters. Unlike DDS, the solution to \eqref{eq:diffpir} is obtained as a closed-form solution.
These hyper-parameters are found through grid search. SR$\times 8$: $\zeta = 0.35, \lambda = 35.0$ / Deblur: $\zeta = 0.3, \lambda = 7.0$ / Inpaint: $\zeta = 1.0 / \lambda = 7.0$.

\begin{table}[!t]
\centering
\resizebox{1.0\textwidth}{!}{%
\begin{tabular}{llllllllllll}
\toprule
{steps} & {$0$} & \multicolumn{3}{c}{$1$} & \multicolumn{3}{c}{$3$} & \multicolumn{3}{c}{$5$} \\
\cmidrule(lr){1-2}
\cmidrule(lr){3-5}
\cmidrule(lr){6-8}
\cmidrule(lr){9-11}
{lr} & {-} & $1e-5$ & $1e-4$ & $1e-3$ & $1e-5$ & $1e-4$ & $1e-3$ & $1e-5$ & $1e-4$ & $1e-3$ \\
\midrule
FID & 61.16 & 60.66 & 59.60 & \textbf{57.61} & 60.11 & 59.34 & 60.19 & 60.02 & \underline{58.59} & 62.67 \\
PSNR & 26.49 & 26.69 & 26.71 & 26.73 & \textbf{26.78} & 26.70 & 26.61 & \underline{26.73} & 26.17 & 26.38 \\
\bottomrule
\end{tabular}
}
\caption{
Robustness to hyper-parameters in prompt-tuning. FFHQ SR$\times 8$ on 256 test images. \textbf{Bold}: best, \underline{underline}: second best.
}
\label{tab:pt_robustness}
\end{table}
\begin{figure}[!thb]
    \centering
    \includegraphics[width=1.0\textwidth]{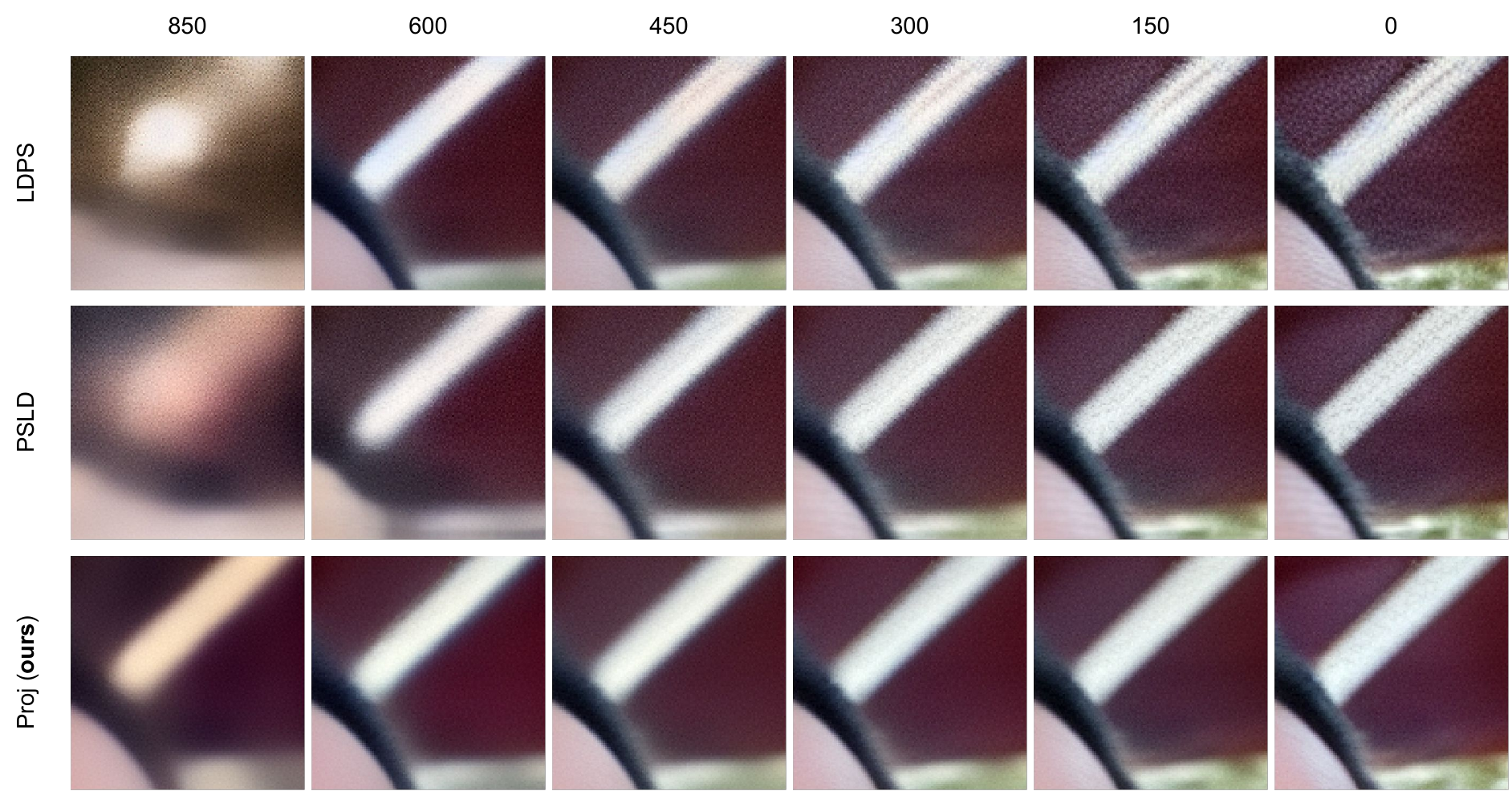}
    \caption{Close-up of the progress of $\Dc(\hat\z_0)$ through time $t$ when solving $\times 8$ SR on FFHQ.}
    \label{fig:progress}
\end{figure}

\section{Efficient implementation in JAX}
\label{sec:implementation_in_jax}

\begin{figure}[h]
\begin{lstlisting}[language=Python]
ones = jnp.ones(x.shape)
_, _AT = jax.vjp(A_funcs.A, ones)
AT = lambda y: _AT(y)[0]
A_funcs.AT = AT
def cg_A(x, cg_lamb):
    return A_funcs.AT(A_funcs.A(x)) + cg_lamb * x
hatx0 = D(hatz0)
cg_y = A_funcs.AT(y) + cg_lamb * hatx0
hatx0, _ = jax.scipy.sparse.linalg.cg(cg_A, cg_y, x0=hatx0)
\end{lstlisting}
\caption{Defining $\Ab^\top$ can be automatically achieved through \texttt{jax.vjp} given that $\Ab$ is differentiable.}
\label{fig:jax_AT}
\end{figure}

In model-based inverse problem solving, having access to efficient computation of the adjoint $\Ab^\top$ is a must. Here, we consider a general case of solving linear inverse problems where the computation of SVD is too costly, and hence one has to define the adjoint operator manually (e.g. computed tomography). Furthermore, for cases such as deblurring from circular convolution, one needs to carefully design the operator, as there are many potential pitfalls (e.g. boundary, size mismatch). These are more often than not the limiting factors of the applicability of the model-based approaches for solving inverse problems. We show in Fig.~\ref{fig:jax_AT} that this can be much alleviated by using \texttt{jax}, as we can implicitly define a transpose operator with reverse-mode automatic differentiation~\citep{baydin2018automatic}. We note this design was also established in~\citep{scico-2022}.

\section{Targetting arbitrary resolution}
\label{sec:highres}

\begin{figure}[!thb]
    \centering
    \begin{tikzpicture}
        \node[anchor=south west,inner sep=0] (image) at (0,0) {\includegraphics[width=1.0\textwidth]{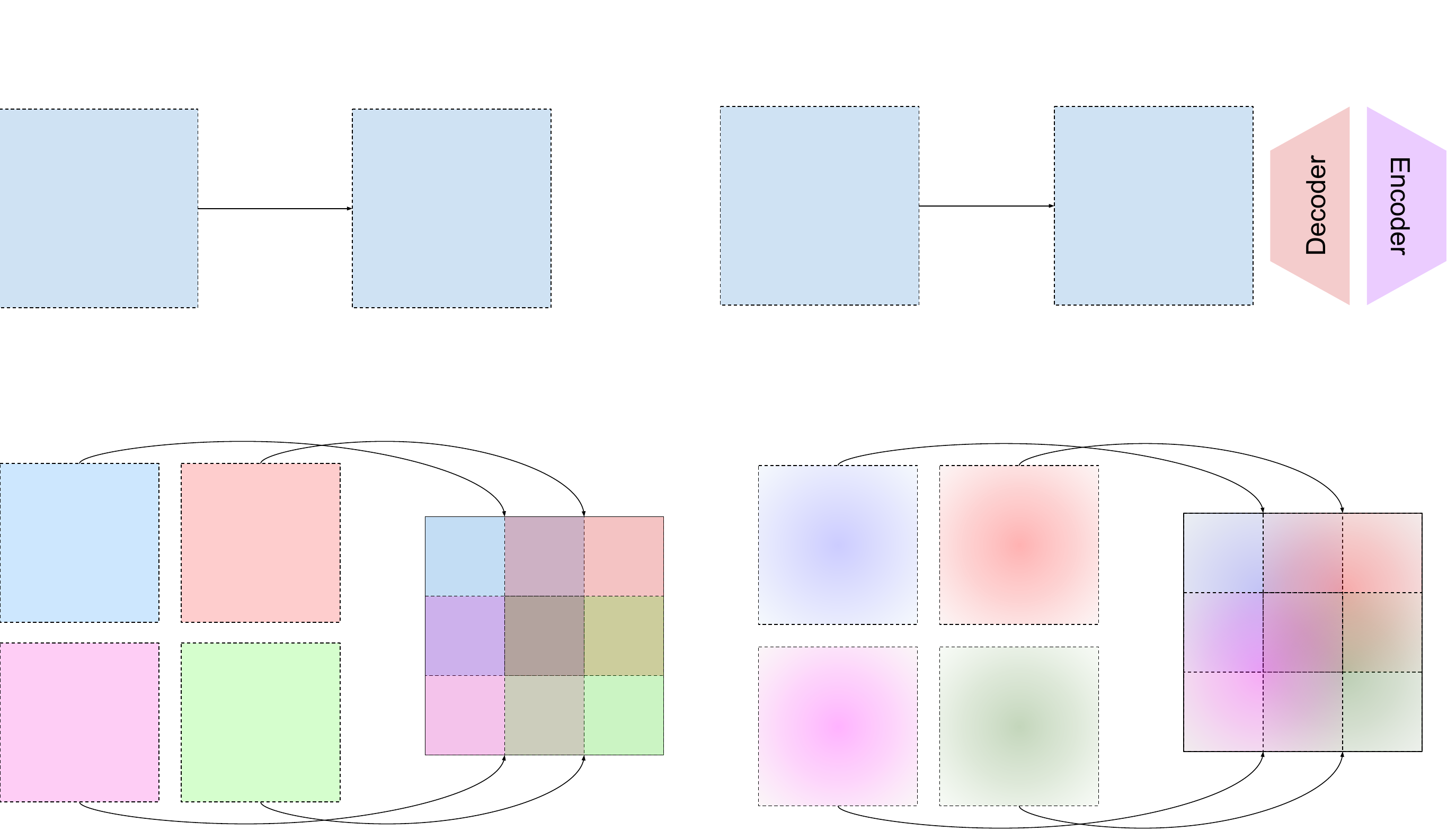}};
        \begin{scope}[x={(image.south east)},y={(image.north west)}]
            \node[text width=4cm, align=left] at (0.1,0.95) {{(a) Vanilla}};
            \node[text width=4cm, align=left] at (0.1,0.55) {{(c) \cite{bar2023multidiffusion}}};
            \node[text width=4cm, align=left] at (0.65,0.95) {{(b) Proposed }};
            \node[text width=4cm, align=left] at (0.65,0.55) {{(d) \cite{jimenez2023mixture}}};
        \end{scope}
    \end{tikzpicture}
    \caption{Method comparison for processing higher resolution images in the latent space.}
    \label{fig:highres_method_overview}
\end{figure}

For SD, using an encoder to convert from the image to the latent space reduces the dimension by $\times 8$. When training SD, the diffusion model that operates on the latent space was trained with 64$\times$64 latents, obtained from 512$\times$512 images. When the image that we wish to restore (or generate) is larger than 512$\times$512, the latents will also be larger than 64$\times$64. In this case, due to the train-test time discrepancy, the results that we get will be suboptimal if one processes the larger latent as a whole (Fig.~\ref{fig:highres_method_overview} (a)). A natural way to counteract this discrepancy is to process the latents in patches\footnote{For all the experiments considered in this paper, we consider 768$\times$768 images (96$\times$96 latents).}. When processing in patches of size 64$\times$64 with stride 32 on both directions, it requires us 4 score function NFEs per denoising step (Fig.~\ref{fig:highres_method_overview} (b),(c)). \cite{bar2023multidiffusion} uniformly weights the overlapping patches, and \cite{jimenez2023mixture} weights the patches with Gaussian weights with variance 0.01. The downside of these methods is that the number NFEs required for inference scales quadratically with the size of the image.

\begin{figure}[!thb]
    \centering
    \includegraphics[width=1.0\textwidth]{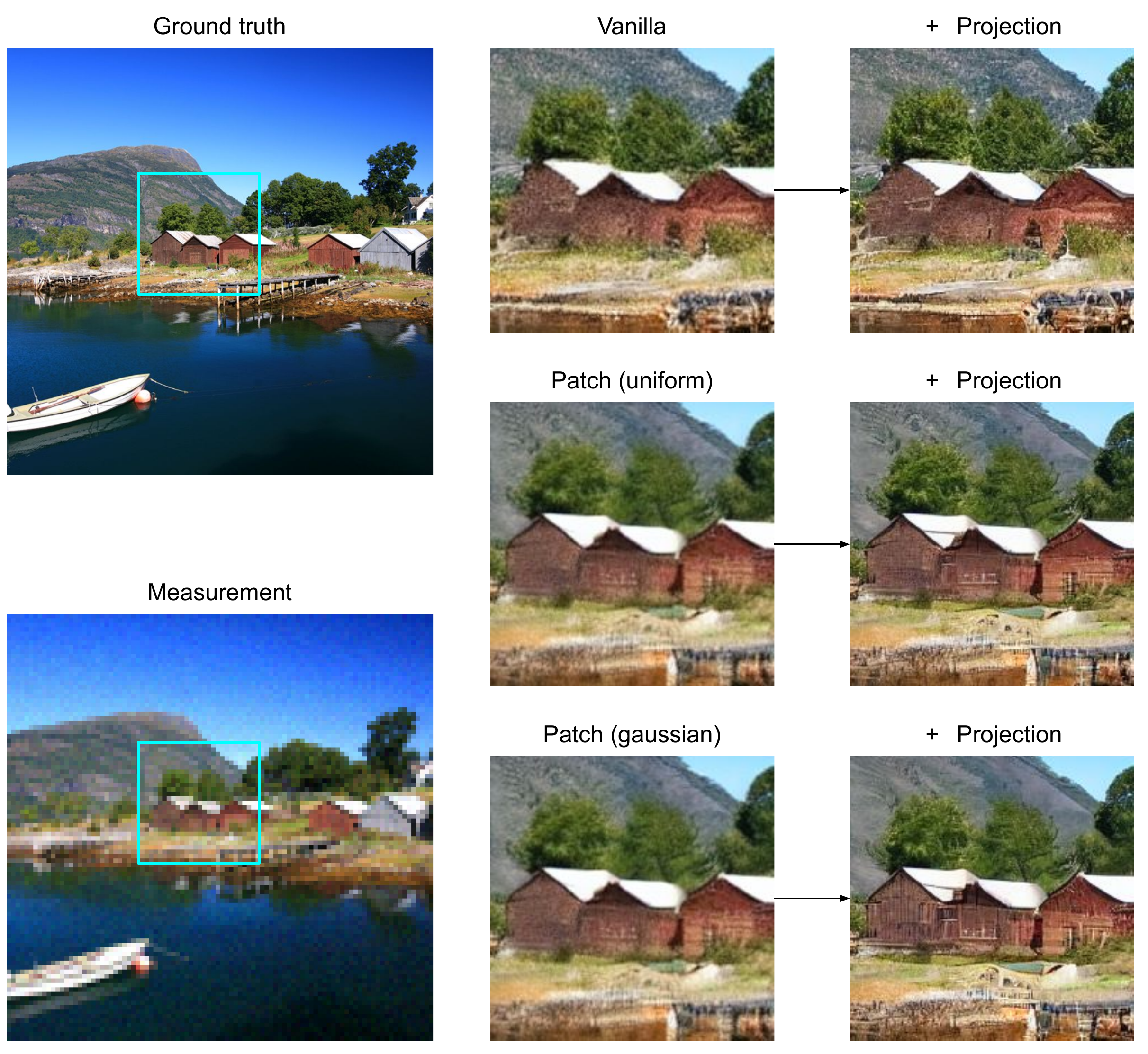}
    \caption{Further results on $\times 8$ SR on DIV2K validation set of 768$\times$768 resolution. Comparison between with and without using our projection approach on various baseline methods.}
    \label{fig:highres_appendix}  
\end{figure}

On the other hand, the proposed method can process the large latent as a whole, as in the ``vanilla'' method, and project this latent to the range space of $\Ec$ by setting $\hat\z'_0 = \Dc(\bm{\Gamma}(\Ec(\hat\z_0)))$ for every few steps. Even though the proposed method is considerably faster than patch-based methods~\citep{bar2023multidiffusion,jimenez2023mixture}, we see that one can achieve a comparable, or superior performance, as presented in Fig.~\ref{fig:highres_exp}. Furthermore, in Fig.~\ref{fig:highres_appendix}, we show that we can use both patching method and the projection method simultaneously, achieving the best results.

\section{Further experimental results}
\label{sec:further_experimental_results}

\begin{figure}[!t]
    \centering
    \includegraphics[width=1.0\textwidth]{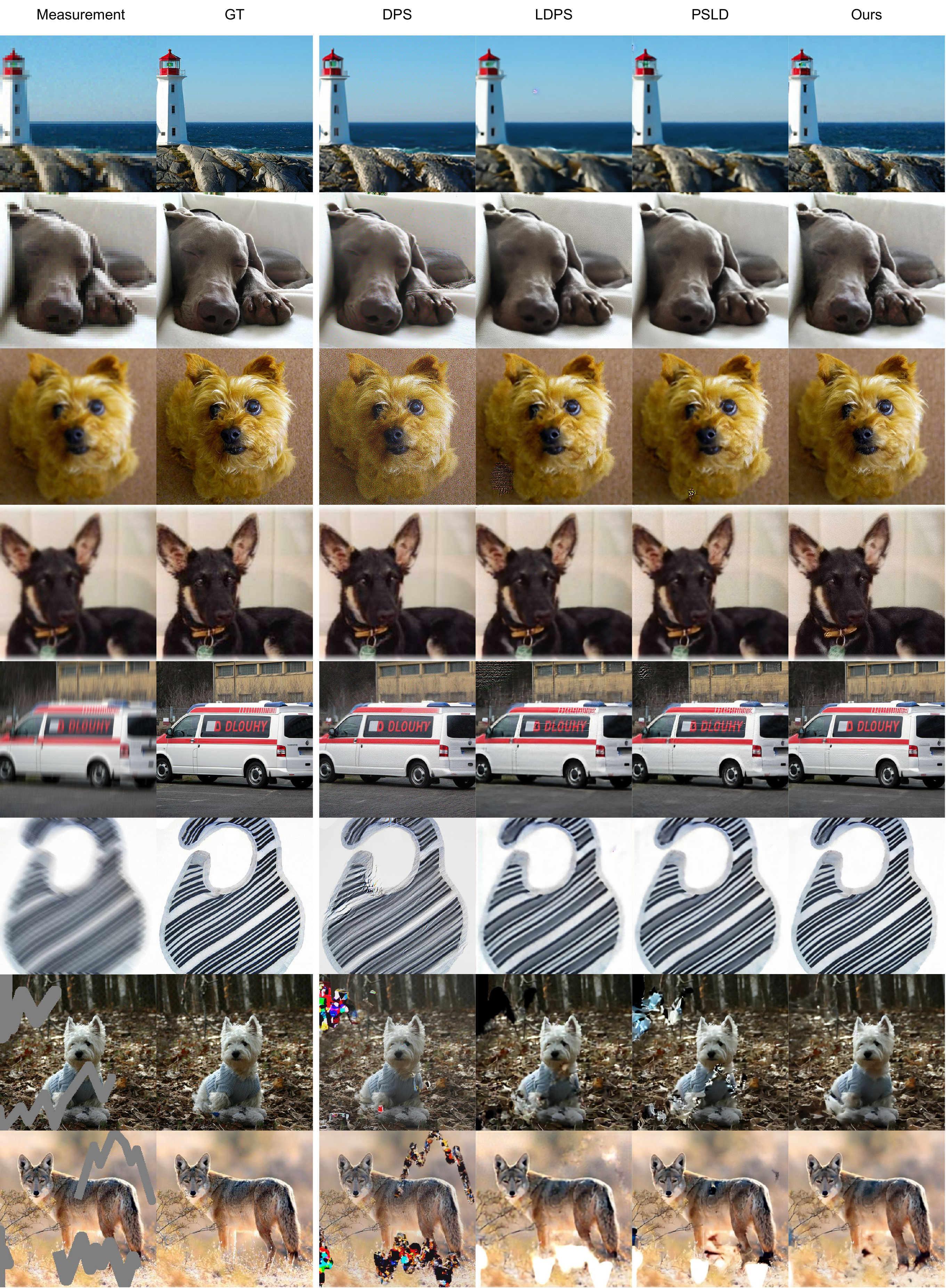}
    \caption{ImageNet restoration results. Row 1-2: SR$\times 8$, row 3-4: gaussian deblurring, row 5-6: motion deblurring, row 7-8: freeform inpainting; All with $\sigma = 0.01$ noise.}
    \label{fig:results_imagenet_appendix}  
\end{figure}

\begin{figure}[!t]
    \centering
    \includegraphics[width=1.0\textwidth]{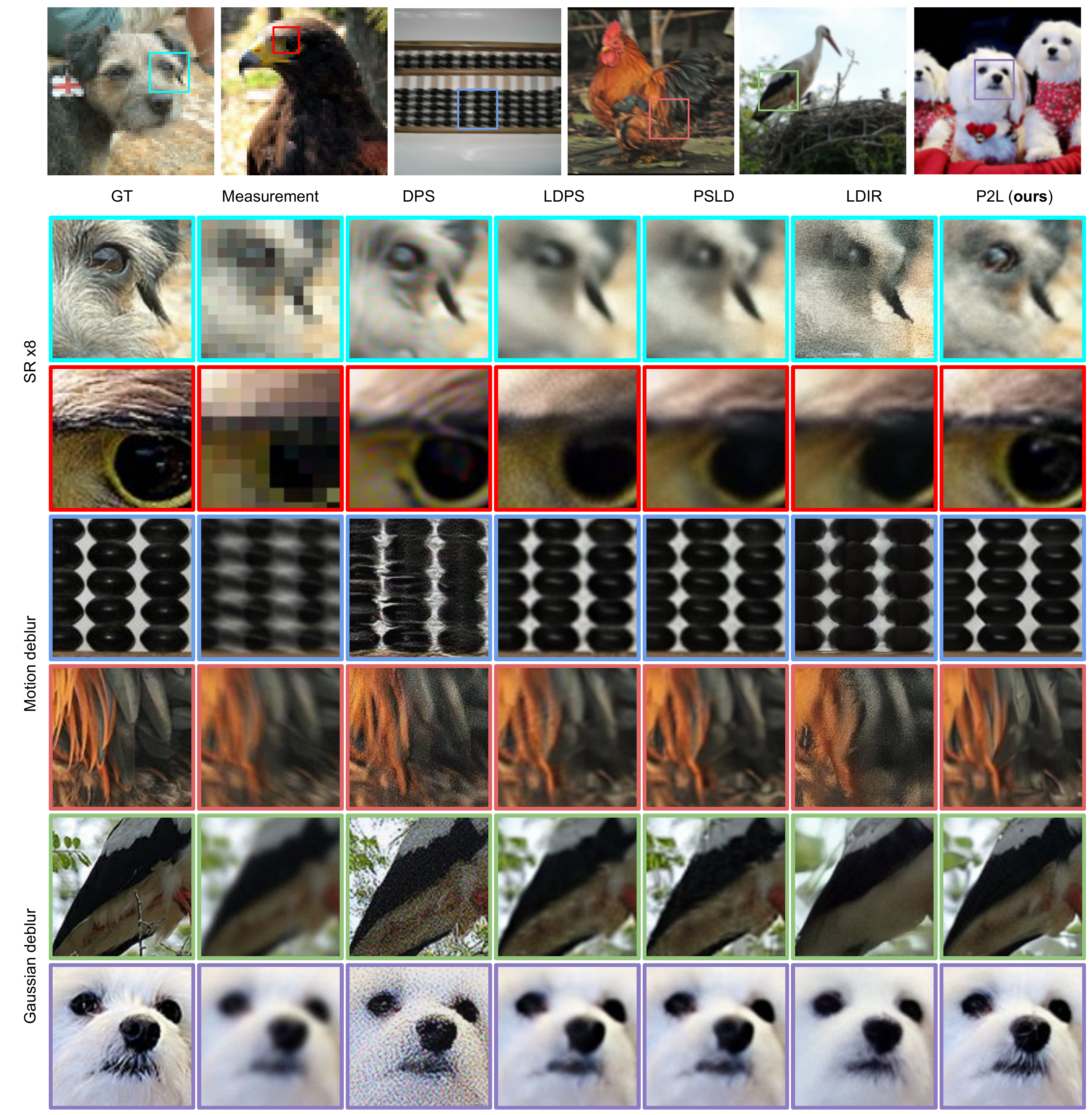}
    \caption{Close-up comparison on diverse inverse problem tasks. Ground truth, measurement, DPS~\citep{chung2023diffusion}, LDPS, PSLD~\citep{rout2023solving}, LDIR~\citep{he2023iterative}, and the proposed method.}
    \label{fig:results_closeup}  
\end{figure}

\end{document}

%% file: packages.tex
\usepackage[utf8]{inputenc} 
\usepackage[T1]{fontenc}    
\usepackage{hyperref}       
\usepackage{url}            
\usepackage{booktabs}       
\usepackage{amsfonts}       
\usepackage{nicefrac}       
\usepackage{microtype}      
\usepackage{xcolor}       
\usepackage{xkcdcolors}
\usepackage{times}
\usepackage{epsfig}
\usepackage{graphicx}
\usepackage{placeins}
\usepackage{multirow}
\usepackage[skip=5pt]{caption}
\usepackage{rotating}
\usepackage{cancel}
\usepackage{setspace}
\usepackage{eso-pic}

\usepackage{bm}
\usepackage{bibunits} 

\usepackage{amssymb,amsthm}
\usepackage{hyperref}
\usepackage{amsmath}
\usepackage{graphicx}
\usepackage{wrapfig,lipsum,booktabs}

\usepackage{epsfig}
\usepackage{tikz}
\usetikzlibrary{spy}
\usepackage{algpseudocode}
\usepackage{algorithm}
\usepackage{mathrsfs}

\usepackage{nicefrac}       
\usepackage{booktabs}       

\usepackage{thmtools,thm-restate}
\usepackage{cleveref}
\usepackage{listings}
\usepackage{lstautogobble}  %
\usepackage{color}          %
\usepackage{zi4}            %
\definecolor{bluekeywords}{rgb}{0.13, 0.13, 1}
\definecolor{greencomments}{rgb}{0, 0.5, 0}
\definecolor{redstrings}{rgb}{0.9, 0, 0}
\definecolor{graynumbers}{rgb}{0.5, 0.5, 0.5}
\usepackage{subcaption}        
\usepackage{siunitx}               
\usepackage[export]{adjustbox}
\usepackage{hyperref}
\usepackage{makecell}
\usepackage{url}
\usepackage{tikz, pgfplots}
\usetikzlibrary{positioning}
\usepackage{capt-of}
\usepackage{diagbox}

\def\eqref#1{(\ref{#1})}

\usepackage{colortbl}



%% file: math_commands.tex
\usepackage{amsmath,amsfonts,bm}

\def\eqref#1{equation~\ref{#1}}

\def\1{\bm{1}}

\def\vv{{\bm{v}}}

\newcommand{\Ib}{{\bm I}}

\DeclareMathAlphabet{\mathsfit}{\encodingdefault}{\sfdefault}{m}{sl}
\SetMathAlphabet{\mathsfit}{bold}{\encodingdefault}{\sfdefault}{bx}{n}

\DeclareMathOperator*{\argmin}{arg\,min}

\newcommand{\cb}{{\boldsymbol c}}

\newcommand{\xb}{{\boldsymbol x}}

\newcommand{\zb}{{\boldsymbol z}}

\newcommand{\gb}{{\boldsymbol g}}

\newcommand{\s}{{\boldsymbol s}}
\newcommand{\n}{{\boldsymbol n}}
\newcommand{\x}{{\boldsymbol x}}
\newcommand{\y}{{\boldsymbol y}}
\newcommand{\z}{{\boldsymbol z}}
\newcommand{\w}{{\boldsymbol w}}
\newcommand{\m}{{\boldsymbol m}}
\newcommand{\g}{{\boldsymbol g}}

\newcommand{\epsilonb}{{\boldsymbol \epsilon}}

\newcommand{\thetab}{{\boldsymbol \theta}}

\newcommand{\Ab}{{\boldsymbol A}}

\newcommand{\Ed}{{\mathbb E}}
\newcommand{\Rd}{{\mathbb R}}

\newcommand{\Ac}{{\mathcal A}}
\newcommand{\Cc}{{\mathcal C}}
\newcommand{\Dc}{{\mathcal D}}
\newcommand{\Ec}{{\mathcal E}}

\newcommand{\Nc}{{\mathcal N}}

\newcommand{\Sc}{{\mathcal S}}

\newcommand{\Lc}{{\mathcal L}}

\newcommand{\code}[1] {\texttt{#1}}

\usetikzlibrary{positioning}
\usepackage{capt-of}
\usepackage{diagbox}

\definecolor{C0}{rgb}{0.121569, 0.466667, 0.705882}
\definecolor{C1}{rgb}{1.000000, 0.498039, 0.054902}
\definecolor{C2}{rgb}{0.172549, 0.627451, 0.172549}
\definecolor{C3}{rgb}{0.839216, 0.152941, 0.156863}
\definecolor{C4}{rgb}{0.580392, 0.403922, 0.741176}
\definecolor{C5}{rgb}{0.549020, 0.337255, 0.294118}
\definecolor{C6}{rgb}{0.890196, 0.466667, 0.760784}
\definecolor{C7}{rgb}{0.498039, 0.498039, 0.498039}
\definecolor{C8}{rgb}{0.737255, 0.741176, 0.133333}
\definecolor{C9}{rgb}{0.090196, 0.745098, 0.811765}
\definecolor{trolleygrey}{rgb}{0.5, 0.5, 0.5}

\definecolor{BrickRed}{rgb}{0.6,0,0}
\definecolor{RoyalBlue}{rgb}{0,0,0.8}
\definecolor{Tdgreen}{rgb}{0,0.4,0.7}
\definecolor{pinegreen}{rgb}{0.0, 0.47, 0.44}
\definecolor{cornellred}{rgb}{0.7, 0.11, 0.11}
\definecolor{cadmiumgreen}{rgb}{0.0, 0.42, 0.24}
\definecolor{spirodiscoball}{rgb}{0.06, 0.75, 0.99}
\definecolor{mylightblue}{rgb}{0.85, 0.90, 0.94}
\definecolor{maroon}{cmyk}{0,0.87,0.68,0.32}

\usepackage{pifont}
\newcommand{\cmark}{\ding{51}}%
\newcommand{\xmark}{\ding{55}}%
\algnewcommand{\LineComment}[1]{\State \(\triangleright\) #1}